\documentclass[letterpaper, 10 pt, conference]{ieeeconf}  % Comment this line out if you need a4paper

\IEEEoverridecommandlockouts                              % This command is only needed if 
\usepackage{booktabs}
\long\def\invis#1{}

\title{\LARGE \bf
EgoExo++: Integrating On-demand Exocentric Visuals with\\ 2.5D Ground Surface Estimation for\\ Interactive Teleoperation of Underwater ROVs
}

\author{Adnan Abdullah$^\star$, Ruo Chen$^{\star}$, Ioannis Rekleitis$^{\diamond}$, and Md Jahidul Islam$^\star$ \\
$^{\star}$RoboPI Laboratory, Dept. of ECE, University of Florida, USA \\
$^{\diamond}$Dept. of ME, University of Delaware, USA
\thanks{\noindent\rule{5cm}{0.5pt}}
\thanks{Accepted for Publication in IJRR, 2026.}
\thanks{Corresponding author: \tt\small adnanabdullah@ufl.edu}%
}

\usepackage{graphicx} % for pdf, bitmapped graphics files
\usepackage{subcaption}

\usepackage{amsmath}
\usepackage{amssymb}

\usepackage{csquotes}

\usepackage{array}
\usepackage{enumerate}
\usepackage{multirow} % tables
\usepackage{float}
\usepackage{paralist}
\usepackage{amsfonts}
\usepackage{longtable}

\usepackage{lipsum}
\usepackage{gensymb}
\newcolumntype{L}[1]{>{\raggedright\let\newline\\\arraybackslash\hspace{0pt}}m{#1}}
\newcolumntype{C}[1]{>{\centering\let\newline\\\arraybackslash\hspace{0pt}}m{#1}}
\newcolumntype{R}[1]{>{\raggedleft\let\newline\\\arraybackslash\hspace{0pt}}m{#1}}

\graphicspath{ {img/} }

\newcommand{\nostarnote}[1]{}
\newcommand{\baad}[1]{} 
\usepackage[size=small]{caption}
\newcommand{\ie}{\textit{i.e.}}
\newcommand{\eg}{\textit{e.g.}}
\newcommand{\etal}{\textit{et al.}} 

\usepackage{hyperref}
\usepackage{color}

\usepackage{cleveref}
\crefformat{section}{\S#2#1#3} 
\crefformat{subsection}{\S#2#1#3}
\crefformat{subsubsection}{\S#2#1#3}

\let\labelindent\relax
\usepackage{enumitem}

\usepackage{makecell}

\usepackage{color}
\usepackage{url}

\newcommand{\extension}[1]{#1}

\newcommand{\rebuttal}[1]{#1}

\newcommand{\finalrev}[1]{#1}

\long\def\invis#1{}

\begin{document}

\maketitle

% \thispagestyle{plain}
% \pagestyle{plain}

%%%%%%%%%%%%%%%%%%%%%%%%%%%%%%%%%%%%%%%%%%%%%%%%%%%%%%%%%%%%%%%%%%%%%%%%%%%%%%%%
\begin{abstract}
Underwater ROVs (Remotely Operated Vehicles) are indispensable for subsea exploration and task execution, yet typical teleoperation engines based on egocentric (first-person) video feeds restrict human operators' field-of-view and limit precise maneuvering in complex, unstructured underwater environments. To address this, we first propose EgoExo, a geometry-driven solution integrated into a visual SLAM pipeline that synthesizes on-demand exocentric (third-person) views from egocentric camera feeds. We further propose EgoExo++, which extends beyond 2D exocentric view synthesis (EgoExo) to augment a piecewise planar 2.5D ground surface estimation on-the-fly. Its anchor-free aerial viewpoint supports ground-relative reasoning, such as clearance and terrain-based navigation marker following. The computations involved are closed-form and rely solely on egocentric views and monocular SLAM estimates, which makes it portable across existing teleoperation engines and robust to varying waterbody characteristics. We validate the geometric accuracy of our approach through extensive experiments of 2-DOF indoor navigation and 6-DOF underwater cave exploration in challenging low-light conditions. To assess operational benefits, we conduct two user studies with simulation and real-world data, each involving $15$ participants, comparing baseline egocentric teleoperation and EgoExo++. Results indicate improved system usability (SUS), reduced perceived workload (NASA-TLX), and significant gains in objective teleoperation performance, including $16\%$ faster missions, $5$-fold reduction in path deviation ratio, and fewer collision events (2 vs. 5 across trials).
% Quantitative metrics confirm the reliability of the rendered exocentric views, while a user study involving $15$ operators demonstrates improved situational awareness, navigation safety, and task efficiency during teleoperation. 
Furthermore, we highlight the role of EgoExo++ augmented visuals in supporting shared autonomy, operator training, and embodied teleoperation. This new interactive approach to ROV teleoperation presents promising opportunities for future research in subsea telerobotics. The source packages for EgoExo++ are available at: \url{https://github.com/uf-robopi/EgoExo}.
\end{abstract}

\section{Introduction}
% \extension{Tentative titles: 
% \begin{enumerate}
%     \item EgoToExo++: Beyond Exocentric Views with 2.5D Ground-Skin Mapping for Interactive Subsea Teleoperation
%     \item EgoToExo++: Dynamic Exocentric and 2.5D Aerial Views for Interactive ROV Teleoperation
% \end{enumerate}
% }

Unmanned submersible vehicles such as ROVs (Remotely Operated Vehicles) play a crucial role in subsea inspection, remote surveillance, and underwater cave exploration~\cite{rumson2021application,wishnak2022new,siegel2023robotic}. They are particularly useful in inspecting deep-water structures and surveying confined spaces that are beyond the reach of human scuba divers~\cite{JoshiAUV2022,buzzacott2009american}. In a typical mission, ROVs are controlled by human operators from a surface vessel, who are responsible for the safe and efficient maneuvering of the vehicle~\cite{konoplin2019development,kennedy2019unknown}. The control consoles for teleoperation typically offer real-time data such as the egocentric video feed, pose, velocity, depth, etc. State-of-the-art (SOTA) ROVs can also include autonomous features for atomic tasks such as hovering~\cite{jin2022hovering}, following navigation guidelines inside underwater caves and overhead structures~\cite{abdullah2023caveseg,yu2023weakly,MohammadiICMLA2023}, object manipulation~\cite{manjunatha2018low,chen2025subsene}, trajectory estimation, etc.

While the subsea industries and agencies such as NOAA and naval defense teams deploy underwater ROVs with high-end cameras, sonars, and IMUs~\cite{wishnak2022new,elor2021catching} -- safe and efficient teleoperation remains a challenge in adverse visibility conditions and around complex or sensitive structures. The typical first-person feeds from an ROV camera provide very limited information in landmark-deprived underwater scenes. The operators on the surface can only see the egocentric view, often without global or peripheral semantic information around the ROV~\cite{lensgraf2023buoyancy,thatipelli2025egocentric}. Although ROVs can use artificial lights to enhance visibility in low-light scenes, their bright lights get reflected and back-scattered by suspended particles directly at the front camera~\cite{yu2023weakly}, creating glare and large blind spots for the operator. Additionally, the semi-autonomous features of ROVs become erroneous without peripheral positioning in such noisy sensing conditions.

\begin{figure*}[t]
\centering
\includegraphics[width=\linewidth]{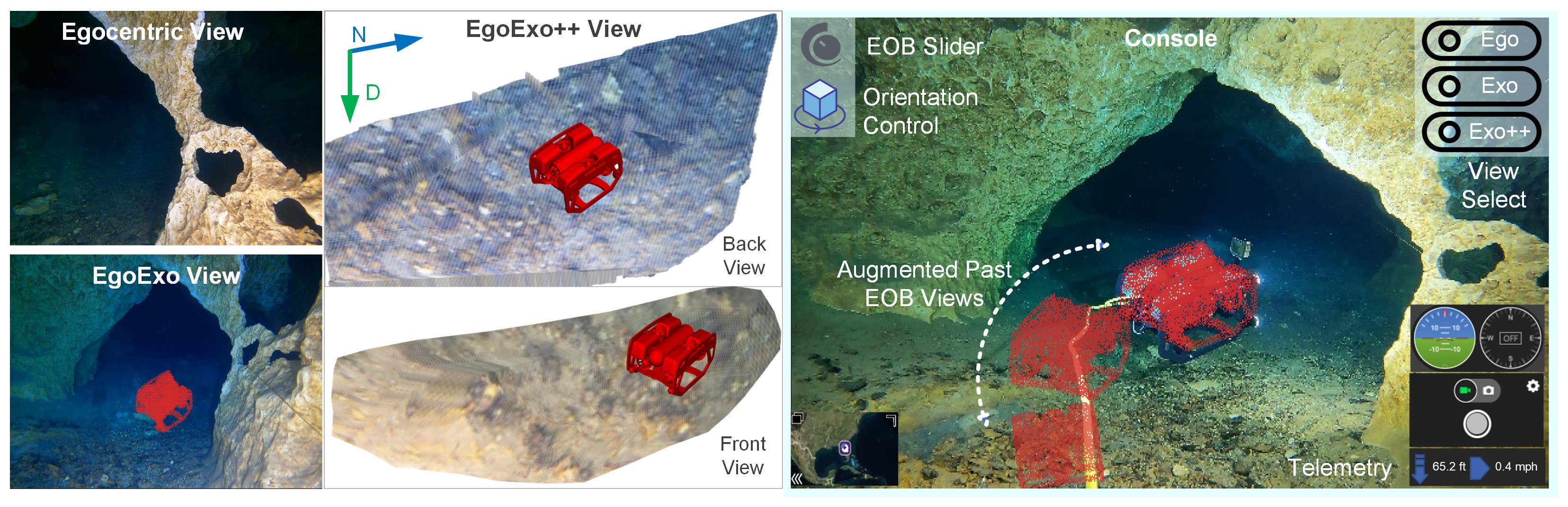}%
\vspace{-3mm}
\caption{\extension{The proposed teleoperation interface is demonstrated for an underwater cave exploration scenario with an ROV. The traditional console interfaces are based on egocentric views (top left), which are limiting and disorienting to a surface operator in noisy low-light conditions. Our \textbf{EgoExo} solution~\cite{abdullah2024ego} offers on-demand exocentric views from a fixed EOB (eye on the back) viewpoint, \ie, third-person views from behind the ROV (bottom left). In \textbf{EgoExo++}, we further integrate dynamic 2.5D exocentric views, with the ROV rendered above a textured ground surface. These interactive view options are integrated into a standard BlueROV2 console (by Blue Robotics Inc.) for a significantly improved teleoperation experience. 
}}
\label{fig:intro}
\vspace{-2mm}
\end{figure*}

In this paper, we address these issues by introducing an AR (augmented reality) inspired ROV teleoperation interface that generates third-person (exocentric) perspectives as well as provides interactive control choices for viewpoint selection. As shown in Figure~\ref{fig:intro}, the proposed console can generate multiple exocentric views from past egocentric images, with a virtual ROV model projected on the images as if it were taken by a \textit{third person} following the robot. \extension{Our early work introduced the idea of \textbf{EOB} (Eye On the Back) visuals~\cite{islam2024eob}, envisioning a single third-person view from immediately behind the ROV to facilitate better teleoperation. Our recent work materialized this idea in \textbf{EgoExo}~\cite{abdullah2024ego}, by formalizing an AR-based framework that generates on-demand exocentric imagery from any EOB viewpoint. It also integrated the feature for geometrically accurate ROV positioning into those views. This work further advances this direction of research by introducing \textbf{EgoExo++}, a dynamic 2.5D exocentric visualization -- analogous to a bird's-eye view in terrestrial contexts -- that offers an interactive and semantically enriched perspective of the environment. 
% \finalrev{EgoExo++ retains the original EgoExo trailing-views and adds the ground-referenced viewpoints, making it a more complete system over EgoExo.}
Importantly, our approach is closed-form and solely geometry-driven, ensuring accuracy and real-time efficiency without reliance on data-driven methods or training biases.
% The proposed \textbf{EgoExo++} system encodes surface-level semantic context to improve operator situational awareness. 
The envisioned interface supports both fore-aft transitions across multiple \textit{EOB views} and an interactive, rotatable $360\degree$ exocentric perspective, enabling a safer and more informed ROV teleoperation.}

\extension{Specifically, we introduce an efficient framework for generating egocentric to exocentric visual perspectives integrated into a visual SLAM system for underwater ROV teleoperation. The base EgoExo algorithm keeps track of the ROV camera poses and exploits a buffer of egocentric views for exocentric view synthesis. We then transform and project a pre-sampled 3D model of the ROV, in the form of a point cloud, into those views to generate realistic augmented visuals with more peripheral information. In parallel, the EgoExo++ pipeline utilizes SLAM-generated feature points to identify ground regions and fuses them into a \rebuttal{piecewise-planar \textit{ground surface}} where pixel colors are transferred from corresponding image regions. \rebuttal{This 2.5D ground reconstruction is particularly meaningful in seabed/structure inspection tasks as well as underwater cave missions where navigation cues such as caveline, arrows, and cookies~\cite{abdullah2023caveseg} are located on or near the floor. } 
%thus, rendering the ground surface improves guided navigation tasks.} 
In our implementation, we employ a temporal \textit{fuse and stack} strategy to preserve the historical ground evidence, while the 3D ROV model is projected on the same spatial context. The resulting 2.5D  perspective enables operators to interact with the scene using dynamic viewpoints in real-time.}
As illustrated in Figure~\ref{fig:intro}, these views provide operators with a globally informed and semantically rich snapshot of the surrounding environment. In addition to supporting interactive viewpoint control, the SLAM backend delivers real-time updates on camera pose and environmental mapping to better assist with atomic tasks such as obstacle avoidance, object following, next-best-view planning, object manipulation, etc~\cite{chen2025subsene,cai2020three,palomeras2019autonomous}.

\rebuttal{We validate the proposed method through a series of analytical, simulation-based, and real-world experiments conducted in both terrestrial and underwater settings. 
In the base EgoExo system, we first quantify the geometric accuracy of view augmentation with a TurtleBot4 ground robot through reprojection error analysis of known reference points in indoor scenes. Then we conduct underwater cave trials with a BlueROV2 at various geographical locations, which pose unique challenges such as low visibility, turbid water conditions, and moving
shadow effects. In such challenging operation scenarios, $15$ human subjects rate the utility of EgoExo visuals using the System Usability Scale (SUS)~\cite{brooke1996sus}; it achieves an average SUS score of $77.5$, indicating the class of \texttt{Good} usability. We further discuss various challenging cases of caveline detection and following in noisy low-light conditions inside multiple underwater caves. These findings establish EgoExo's trailing views as a useful teleoperation aid, and also highlight the critical role of maintaining altitude clearance and detecting on-ground navigation markers, motivating the semantic ground representation of EgoExo++.}

\vspace{1mm}
\noindent
\textbf{EgoExo to EgoExo++: new capabilities.} Our base EgoExo framework leverages the SLAM-estimated 6-DoF pose to synthesize a trailing third-person view by projecting the ROV into a past egocentric image within a global 3D frame. %However, this global world remains \textit{semantically empty}, lacking any representation of the surrounding environment. 
EgoExo++ advances beyond this ``pose only trajectory'' visualization to a semantically enriched global representation by introducing two new capabilities:
\begin{enumerate} [label={(\arabic*)},nolistsep,leftmargin=*] 
    \item EgoExo++ transforms the \textit{semantically empty} world representation of EgoExo into a 2.5D ground-referenced altitude map. It leverages the already available sparse SLAM features to identify and reconstruct locally observable ground patches from each local view, which are then temporally fused in the global frame.
    \item It advances the visualization from a trajectory-anchored, \textit{EOB} perspective to an anchor-free aerial viewpoint, allowing operators to view the ROV from arbitrary vantage points above the reconstructed ground surface and to reason more effectively about altitude clearance and navigation cues.
\end{enumerate}

\vspace{1mm}
\noindent
\textbf{Extended evaluation of EgoExo++.} To demonstrate the new capabilities, we present a detailed technical pipeline for ground surface reconstruction, along with comprehensive experiments. First, we perform geometric analyses to quantify ground segmentation reliability, including plane fitting error, inlier fraction, plane normal consistency, and altitude drift. We then conduct a new user study involving $15$ participants, where users teleoperate an ROV in a Gazebo-based underwater industrial facility~\cite{abdullah2025active} to follow pipelines and detect target markers. We report both subjective workload (NASA-TLX~\cite{hart1988development}) and objective metrics, including mission completion time, path deviation, and collision count. We also conduct a supplementary counterbalanced evaluation in which the experiment order is reversed for a subset of participants to assess potential order bias between the baseline and EgoExo++ trials. The results show consistent performance trends, indicating the benefit of our system.

Notably, EgoExo++ reduces perceived workload while achieving approximately $16\%$ faster mission completion, $5\times$ lower Path Deviation Ratio (PDR), and fewer collisions compared to the baseline (egocentric only) teleoperation system. Together, these additional evaluations demonstrate both technical soundness and operational relevance of the proposed framework. Finally, we discuss the broader potential of the EgoExo++ paradigm for shared autonomy and digital twin-based training, as well as its dependency on SLAM and the practical limitations that arise in challenging underwater environments.

\section{Background and Related Work}

\subsection{Third-Person Views for ROV Teleoperation}
% \vspace{-0.5mm}
A common issue reported by ROV operators is that using a remote vision platform for teleoperation is like looking through a ``\emph{soda straw}"~\cite{woods2004humanrobotcoordination,islam2024eob}. This is because the typical ROV controller interfaces are based on egocentric \textit{first person camera} views -- which provide no peripheral vision, resulting in significantly reduced situational awareness~\cite{casper2003wtc,zollmann2014flyar}. Researchers have explored both fixed~\cite{ferland2009egocentric,lager2018remote} and dynamic~\cite{nguyen2001virtual,okura2013teleoperation} viewpoint augmentation methods in contemporary human-machine interface study~\cite{abdullah2024human,xia2022virtual}.

Two primary approaches are used for generating exocentric views in unmanned ground and aerial vehicles. The first leverages external cameras to capture the vehicle's motion from a distance; examples include fixed ground cameras~\cite{jangir2022thirdpersonmanip}, UAV-mounted overhead views~\cite{gawel2018aerial,saakes2013flyingcameras,inoue2023birdviewar,erat2018drone}, elevated on-robot mounts~\cite{shiroma2004study}, camera-equipped follower ROVs~\cite{nagatani2011quince}, and fisheye lenses for top-down perspectives~\cite{sato2013spatiotemporal,hing2010chaseview}. The second method utilizes additional onboard sensors, such as LiDAR (Light Detection and Ranging), to generate a point cloud of the surrounding environment~\cite{ferland2009egocentric,lager2018remote} and use it to create an augmented/virtual reality for interfacing and teleoperation~\cite{livatino2021intuitiverobotteleop,hing2010chaseview,thomason2019comparison,xia2023sensory}.

% A similar approach to our method can be found in ~\cite{sugimoto2005timefollower}

Adapting the aforementioned methods from terrestrial or aerial domains to underwater environments presents inherent challenges. %Deep underwater environments have significantly different pressure and temperature from room condition. They often suffer from poor visibility due to water turbidity and low light conditions. Hence, 
Firstly, sending diver-robot teams~\cite{islam2021robot} is not always an option in complex deep-water missions -- which are the majority of use cases for ROVs. Secondly, UGVs that utilize past egocentric views~\cite{ito2008teleoperation,murata2014teleoperation} primarily rely on GPS-based localization that does not apply to GPS-denied underwater environments. Unlike underwater ROVs, ground vehicles generally operate on a 2D plane with limited pitch and roll variations over rough terrain~\cite{yoon2025learning}. Thirdly, installing an external visual system requires significant hardware modifications, \eg, they need to be rugged and pressure-sealed, recalibrated for buoyancy and motion dynamics, and tether integration for high-speed exocentric data transfer. Even with all the structural modifications, an external camera will provide a single additional third-person perspective.

\extension{A range of AR/VR-based teleoperation systems have been developed to enhance operator immersion and augment visual feedback for subsea tasks such as object grasping and manipulation~\cite{bruno2018augmented,xia2022virtual,chen2025subsene,girbes2020haptic}, inspection~\cite{blow2025detection,zhou2023embodied}, and navigation~\cite{xia2023visual}. These systems commonly support third-person perspectives by embedding the operator within an extended reality environment that incorporates a digital twin of the ROV~\cite{xia2023rov} and reconstructs the surrounding scene using 3D models. While such immersive interfaces improve situational awareness and control, they often require extensive sensory augmentation (\eg, visual, auditory, haptic) at both the ROV and operator ends~\cite{chen2025subsene,zhou2023embodied,xia2023sensory}, which increases hardware demands and complicates real-time deployment.}

% We attempt to address these issues by synthesizing geometrically accurate exocentric views into a monocular SLAM pipeline given a dynamic viewpoint. 

% \JI{Is this background updated?? So many good papers are missing!!!! This was the reason for IROS rejection, just FYI}

% \JI{For instance, none of Eric's papers are here; I cited one myself. The survey had other papers as far as I remember. Please UPDATE THIS!!!}

\subsection{2.5D Exocentric View Generation}
Generating 2.5D/3D third-person views from front-facing camera is critical for scene understanding, both for human teleoperators and for autonomous vehicles. The challenges lie in extreme viewpoint shift and lack of direct depth cues from monocular inputs. Recent efforts for 2.5D view synthesis can be categorized into two main areas: homography-based geometric projections~\cite{wang2019monocular,abbas2019geometric} and generative models using encoder-decoder, adversarial, or transformer-based learning~\cite{li2024bevformer,luo2024intention}.

\extension{The geometry-guided CNN proposed by~\cite{abbas2019geometric} warps frontal images to the top view using a fitted homography matrix. While efficient for structured environments, their approach is limited to the flat-ground assumption and struggles with non-planar surfaces. Zhu~\etal~\cite{zhu2018generative} introduce an intermediate homography view from generative adversarial network (GAN) to reduce the difficulty of pure geometric transformation. Transformer models such as BEVFormer~\cite{li2024bevformer} and BEVDepth~\cite{li2023bevdepth} integrate temporal or multi-view cues to improve realism in synthesized views at a cost of high computation. Other learning-based approaches fuse multiple camera views or additional sensors (\eg, LiDAR) to generate semantic aerial views~\cite{reiher2020sim2real,samani2023f2bev}, diverging from monocular egocentric setups. Unlike these data-driven approaches, we propose a lightweight geometric solution, integrated into a visual SLAM pipeline that offers real-time, interactive third-person perspectives without relying on multi-modal sensory augmentation or additional hardware.}

\begin{figure*}[t]
\centering
    \includegraphics[width=\linewidth]{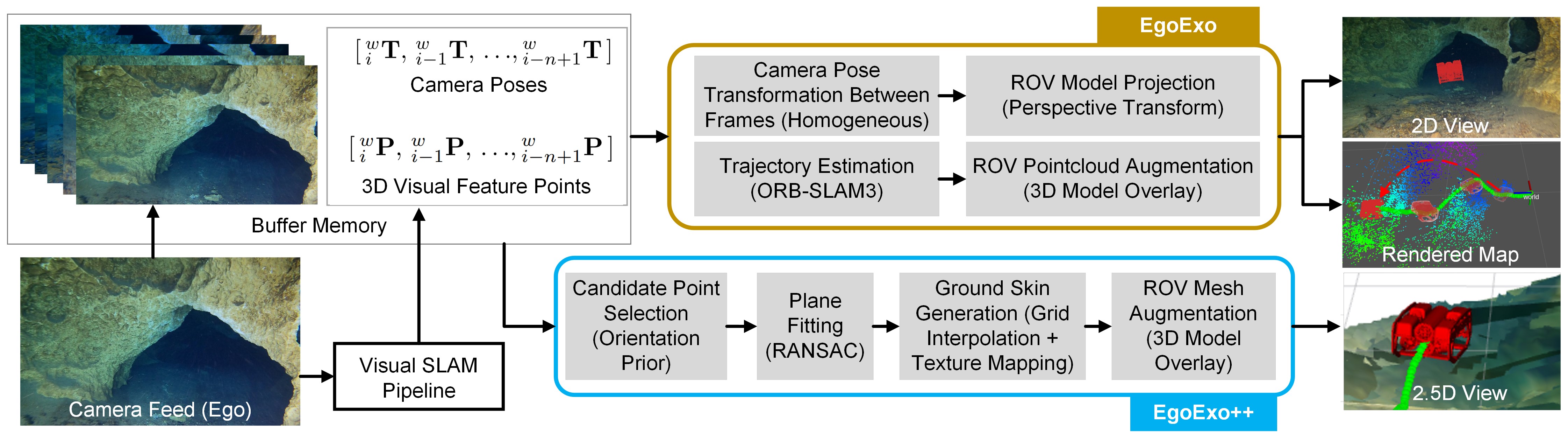}%
    \vspace{-1.5mm}
    \caption{\extension{The computational pipeline is shown. From historical egocentric views and SLAM-derived poses, EgoExo computes a 2D exocentric image by applying pose geometry to project the ROV model; a sparse map of the environment is also constructed using SLAM-derived feature points. EgoExo++ reuses the feature points to fit a ground plane via RANSAC, then generates a textured 2.5D ground surface, and augments the ROV mesh to produce interactive exocentric views. }}
    %The proposed EgoExo++ problem formulation is shown. The idea of EgoExo is to generate EOB (Eye On the Back) views with augmented ROV's pose for more informative third-person perspectives. The EgoExo++ extends from 2D image representations to 2.5D interactive views to facilitate safer and more efficient ROV teleoperation.
\label{fig:models}
\vspace{-3mm}
\end{figure*}

%needs to be more precise, rugged and pressure-sealed to function in such harsh conditions. Such systems also demand for additional space and high-speed data cables to transmit video feed in real-time, which add extra complexity and cost to the platform design. Sending a group of human divers to continuously follow an ROV is too labor-intensive and extremely risky, particularly in confined spaces. A second ROV as follower is not a feasible solution too, since it would be challenging for the second teleoperator to mimic the preceding ROV's motion in 3D space with only egocentric vision available. Furthermore, generating an augmented 3D map of the environment using the limited power and hardware resources onboard ROVs is computationally too demanding.  
%as well as increases the risk of entanglement with nearby structures. On the other hand, ROVs' onboard sensors usually have limited computation ability due to hardware and power restriction.  and  
%To address these issues, we propose to develop an improved TeleOp interface that will use the available egocentric data to generate an augmented peripheral view and provide a safer and smoother teleoperation experience.

\section{EgoExo++: Problem Formulation}
We formulate the EgoExo++ problem as a 3D geometric algorithm that involves generating an on-demand EOB view, reconstructing the ground surface, and then projecting the ROV model both on 2D and 2.5D context for augmented rendering of the scene; see Figure~\ref{fig:models}. The proposed method has the following computational components.
%The proposed method comprises three key components. First, it takes a series of egocentric images and corresponding camera poses obtained from SLAM algorithm, and holds them in a temporary memory. Then a geometric transformation takes place from a \textit{current} frame to a preceding user-defined \textit{reference} frame, where the ROV model is projected accordingly. Concurrently, the method renders the ROV model onto a 3D map using most recent pose information.

\subsection{Curating ROV Pose and Image Buffer}
% \vspace{-0.5mm}
A monocular SLAM algorithm such as ORB-SLAM3~\cite{ORBSLAM3_TRO} provides a continuous solution for estimating and tracking camera poses from a sequence of monocular images. We use an ORB-SLAM3-based framework to obtain camera poses of each keyframe location to eventually construct the trajectory map of the teleoperated robot. In our implementation, the SLAM pipeline initiates the trajectory estimation process by building a pose buffer of length $n$:
${}^{w}\mathbf{T} \triangleq [\,{}^{w}_{i}\mathbf{T},\, {}^{w}_{i-1}\mathbf{T},\, .\,.\,., {}^{w}_{i-n+1}\mathbf{T}\,]$, where, ${}^{w}_{i}\mathbf{T} = [ {}^{w}_{i}\mathbf{R}_{3\times3} \,|\, {}^{w}_{i}\mathbf{t}_{3\times1}]$ denotes camera pose at instance $i$ in global ($world$) frame of reference. The corresponding raw egocentric views $\mathbf{I}$ for each instance are also stored in a queue 
$\mathbf{I} \triangleq [\,\mathbf{I}_i,\, \mathbf{I}_{i-1},\, \cdots \, \mathbf{I}_{i-n+1}\,]$. These memory buffers are updated instantaneously as the robot's pose changes during teleoperation. We use an empirically tuned threshold to trigger an update only when the pose change is significant to avoid unnecessary updates (when the robot is static). 
%Specifically, the queue is updated with each newly available pose and image if the rotation and/or translation from immediate previous instance exceeds a predefined threshold.  

\subsection{Generating 2D Exo Image}
% \vspace{-0.5mm}
Given the pose memory ${}^{w}\mathbf{T}$ and egocentric views $\mathbf{I}$, we formulate the EgoExo problem of estimating an exocentric view from a reference location $r$, looking toward the robot's current location $c$, where $r,c \in [i-n+1,\, i]$ and $r < c$. Typically, $c$ is set to $i$ (most recent available frame), and $r$ remains a free variable with $n$ known samples in memory -- to mimic the EOB viewpoint generation.  

We use the ROV point cloud model $\mathbf{P}_{rov}$ of size $3\times m$ as prior. These $m$ points are transformed from current camera pose ${}^{w}_{c}\mathbf{T}$ to reference camera pose ${}^{w}_{r}\mathbf{T}$ using:
\begin{equation}
    \mathbf{\tilde{P}}_{rov} = ({}^{w}_{r}\mathbf{R}^{-1}\, {}^{w}_{c}\mathbf{R})\cdot \mathbf{P}_{rov} + ({}^{w}_{c}\mathbf{t} - {}^{w}_{r}\mathbf{t}),
\end{equation}
where $[ {}^{w}_{c}\mathbf{R} \,|\, {}^{w}_{c}\mathbf{t} ]$ and $[ {}^{w}_{r}\mathbf{R} \,|\, {}^{w}_{r}\mathbf{t}]$ represent the ROV pose for current and reference (target) location in world coordinate, respectively. The transformed point cloud $\mathbf{\tilde{P}}_{rov}$ is then projected onto the target image plane by using camera intrinsics $\mathbf{K}$ as:
\begin{equation}
\begin{bmatrix}
\mathbf{u} &
\mathbf{v} &
\mathbf{1_{m \times 1}}
\end{bmatrix}^T
= \lambda_{1}\, \mathbf{K}\cdot \mathbf{\tilde{P}}_{rov}.
\end{equation}
Here, $\mathbf{u}$ and $\mathbf{v}$ vectors denote the pixel locations ($u$,$v$) on image $\mathbf{I}_{r}$ for projection; $\lambda_{1}$ is the scale.

\subsection{Generating 2.5D Exo Views}\label{subsec:2.5D_exo_views}
% Talk about- feature point selection, ground plane estimation, homographic projection of ground points, ground skin synthesis with voxel decimation and Delaunay triangulation, rolling buffer of ground skin
In EgoExo++, we reuse the SLAM-generated visual features to estimate the ground surface and synthesize a lightweight terrain-aware 2.5D perspective. \rebuttal{We focus on reconstructing only the ground surface rather than the full volumetric scene since the ground structure provides altitude awareness and a stable spatial anchor for teleoperation. Moreover, navigation markers in underwater caves, such as cavelines and arrows, are typically located on the ground, making ground reconstruction more relevant for guided navigation.}
The process involves four stages: (i) selecting candidate feature points for the ground surface, (ii) fitting the ground plane, (iii) translating texture from image pixels to the estimated surface, and (iv) fusing multiple frames over time for real-time visualization.

Due to the lack of horizon line in open water settings and the uneven geometry of confined underwater spaces (\eg, caves), we incorporate geometric priors based on the camera orientation to initialize the ground region estimation. In the nominal case with zero pitch and roll, the ground remains within the bottom half of the image, separated by a horizontal line at $v=H/2$ (where $H$ is the image height). 
As the camera pitches downward, this line shifts upward, since a larger portion of the ground comes within the camera's FOV, and vice versa. A camera roll rotates this dividing line on the image plane accordingly. 
By computing this orientation-adjusted imaginary horizon from the known camera pose, we restrict candidates to points that fall within the ``ground side'' of the image. This prior ensures that no 3D point projecting above the horizon (\eg, from cave walls and ceiling) is selected as ground.

\extension{
Let ${}^{w}\mathbf{P} = \{\mathbf{p}_j \in \mathbb{R}^3\}_{j=1}^J$ be the set of SLAM feature points in the world frame, associated with camera pose ${}^w_i \mathbf{T}$ at time instance $i$. After imposing the geometric prior and pre-selecting candidate points, we fit a plane $\pi: \mathbf{n}^\top \mathbf{x} + d = 0$ via RANSAC~\cite{fischler1981random}:
\begin{equation}
\label{eq:ransac_plane}
\min_{\mathbf{n},d} \;\sum_{j} \rho\!\left(\big|\mathbf{n}^\top \mathbf{p}_j + d\big|\right),
\end{equation}
where $\rho(\cdot)$ is an inlier loss with threshold $\tau$. To enforce stability, we apply a prior that constrains the plane normal $\mathbf{n}$ within an angle $\pm\theta_{\max}$ of the expected vertical direction ($-y$ in camera frame). Given the plane $\pi$ and a reference anchor $\mathbf{x}_0$ (closest point from camera center to $\pi$), we define an orthonormal basis $\{\mathbf{e}_u,\mathbf{e}_v,\mathbf{n}\}$ on the plane. Each 3D point is expressed in local coordinates as:
\begin{equation}
\begin{bmatrix}
u_j & v_j & h_j
\end{bmatrix}
= (\mathbf{p}_j - \mathbf{x}_0)^\top 
\begin{bmatrix}
\mathbf{e}_u & \mathbf{e}_v & \mathbf{n}
\end{bmatrix}.
\end{equation}
A rectangular grid $(\xi,\eta)$ is constructed on the ground plane, and the sparse heights $\{h_j\}$ are interpolated to obtain a smooth elevation field $h(\xi,\eta)$. Each grid vertex 
\begin{equation}
\mathbf{q}(\xi,\eta) = \mathbf{x}_0 + \xi\,\mathbf{e}_u + \eta\,\mathbf{e}_v + h(\xi,\eta)\,\mathbf{n}
\end{equation}
is then reprojected to the image using intrinsics $\mathbf{K}$:
\begin{equation}
\begin{bmatrix}
u' & v' & 1
\end{bmatrix}
= \lambda_2\, \mathbf{K}\,{}^w_i\mathbf{T}^{-1}\,\,\mathbf{q}(\xi,\eta).
\end{equation}
Image colors $\mathbf{I}(u',v')$ are sampled (bilinear interpolation) to texture the grid, producing a dense 2.5D ground surface. To extend the ground beyond a single camera frame, all historical ground patches are accumulated in the global frame \finalrev{while maintaining local plane normals; we do not assume a globally flat ground plane}. 
% Each frame contributes a colored ground patch in its world-aligned position.
Patches are merged using voxel decimation and Delaunay triangulation~\cite{lee1980two} to avoid redundancy while preserving continuity. The fused mesh forms a 2.5D exocentric perspective with \finalrev{natural uneven terrain variation} and realistic coloring consistent with the egocentric imagery.
}

% \vspace{-2mm}
\subsection{ROV Model Rendering and Scene Update}
While the SLAM system constructs a sparse map of the surroundings, the proposed algorithm simultaneously renders the 3D ROV point cloud (or mesh for EgoExo++) on the same spatial context. The ROV points $\mathbf{P}_{rov}$ are transformed to the current camera location and projected based on the relative pose information ${}^{w}_{i}\mathbf{T}$ as follows: 
\begin{equation}
    \mathbf{\tilde{P}}_{map} = \lambda_{3}\, {}^{w}_{i}\mathbf{R}\cdot \mathbf{P}_{rov} + {}^{w}_{i}\mathbf{t}.
\label{eq3}
\end{equation}
Here, $\lambda_{3}$ is the scaling factor for the ROV model. Note that our mapping and projection method is up to scale, like all monocular SLAM-based systems~\cite{kazerouni2022survey,macario2022comprehensive}. While the scale can be resolved with additional sensor fusion, the augmented visuals of Eq.~\ref{eq3} are sufficient for teleoperation.

%; this number is carefully chosen to optimize the balance between computation time and the accurate representation of the vehicle's structure. The pointclouds are normalized between $-1$ to $1$; however the actual scale is later taken care of by the scale parameters $\lambda_1$ and $\lambda_2$ for the rendered exocentric view and 3D map, respectively. Three different cameras are utilized to collect visual data from a wide range of scenarios. A GoPro10 camera is mounted on BlueROV2 to collect data from underwater caves. On the other hand, the terrestrial data is collected with a USB camera mounted on TurtleBot4. Additional indoor data is gathered using a cellphone camera. Camera calibration is performed for each camera using checkerboard patterns; and their respective intrinsic and distortion parameters are utilized during both SLAM implementation and the Ego-to-Exo method.

% \begin{figure}[h]
% \centering
% \includegraphics[width=\linewidth]{imgs/reprojection_visual.png}%
% \vspace{-2mm}
% \caption{An office setup for quantitative evaluation with checkerboard reference is shown. The $f$ numbers indicate the \textit{EOB distance} from current to the reference frame used for the EgoExo view generation. As the method attempts to project the robot further back into the past, it results in higher reprojection errors % for checkerboard corners; and the robot's representation becomes proportionately unrealistic in the generated EgoExo view.
% }
% \label{fig:reprojection}
% \vspace{-3mm}
% \end{figure}

\section{Implementation \& Evaluation}
\subsection{Implementation Details}
\vspace{-0.5mm}
The framework is implemented using ROS Noetic in an Ubuntu $20.04$ environment, running on an Intel Core i9 processor with $16$\,GB of RAM. A ROS node for ORB-SLAM3 is integrated as the monocular SLAM backbone. Note that we adopt the North-East-Down (NED) frame convention used by~\cite{manderson2016texture}, which is local to the SLAM origin (not aligned with Earth's North/East). 
% \JI{Following someone's paper or convention? then cite. If not seems okay (although some core SLAM people may not like it)}
% \Adnan{Found a paper from Greg that mentions local NED frame}
The scaling parameters $\lambda_1$, $\lambda_2$, and $\lambda_3$ are empirically tuned once for each test sequence according to the scale of the map \finalrev{and the approximate physical dimensions of the ROV to ensure visually realistic projections. Once chosen, the scale parameters remain fixed
throughout the mission and do not require retuning during operation}.

\vspace{1mm}
\noindent
\textbf{EgoExo view augmentation}. We maintain a buffer of past egocentric frames with a queue size of $n = 100$; the frame separation threshold is set to $0.001$ units (up to scale). The ROV point clouds are generated by sampling 3D mesh models of BlueROV2 and TurtleBot4; $10,000$ points are sampled for each model.

\vspace{1mm}
\noindent
\textbf{EgoExo++ ground reconstruction}. We extract the ground patch from each egocentric image using a RANSAC-based plane estimation process. We use a $0.10$\,unit point-to-plane inlier threshold, up to $2000$ iterations, and require at least $50$ inliers, with early termination if more than $80\%$ of candidates are explained. The fitted plane normal in camera coordinates is restricted to stay within $60\degree$ of the negative camera y-axis (downward). All points within a distance band of $0.20$\,unit around the fitted plane, that are (i) in front of the camera, (ii) inside the image, (iii) in the bottom band, and (iv) geometrically below the camera are finally labeled as ground. 

We then express these inliers in a local plane frame and rasterize them onto a regular 2D grid with cell size $0.01$\,unit. Heights on this grid are obtained with a \texttt{LinearNDInterpolator} over the inlier samples, with a \texttt{NearestNDInterpolator} used to fill holes where the linear interpolation is undefined. The resulting dense 3D surface is triangulated as a regular mesh (two triangles per grid cell) provided that all three vertices back-project to valid image pixels inside the convex hull of the ground inliers; for efficiency and decimation, we cap the number of triangles per frame at $20000$.

%\JI{Are these algorithmic details or implementation details? Did they want some reference to Fig2 and a discsussion connecting important equations?}
%\Adnan{This para addresses Qs 2.2. Reviewer asked for the detailed parameters for ground segmentation. It is a mix of criteria and numbers. So I moved it here in Implementation Details.}

\subsection{Proof of Concept: 2D Indoor Navigation}

\begin{figure}[t]
\vspace{1mm}
\centering
\begin{subfigure}[]{\linewidth}
\includegraphics[width=\linewidth]{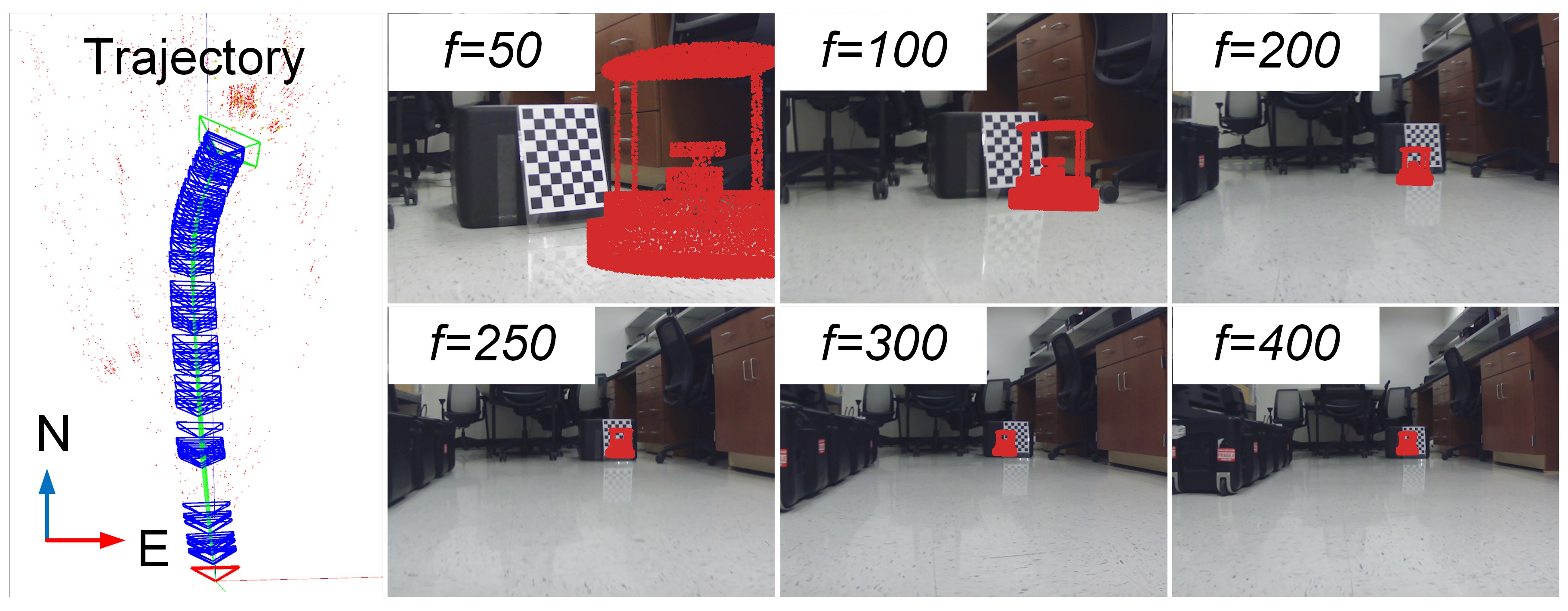}%
\vspace{-2mm}
\caption{The TurtleBot4 trajectory during teleoperation is shown; here, the $f$ numbers indicate the \textit{EOB distance} from current to reference frame used for the generated EgoExo views.}
\label{fig:reprojection_a}
\end{subfigure}

\vspace{1mm}
\begin{subfigure}[]{\linewidth}
\includegraphics[width=\linewidth]
{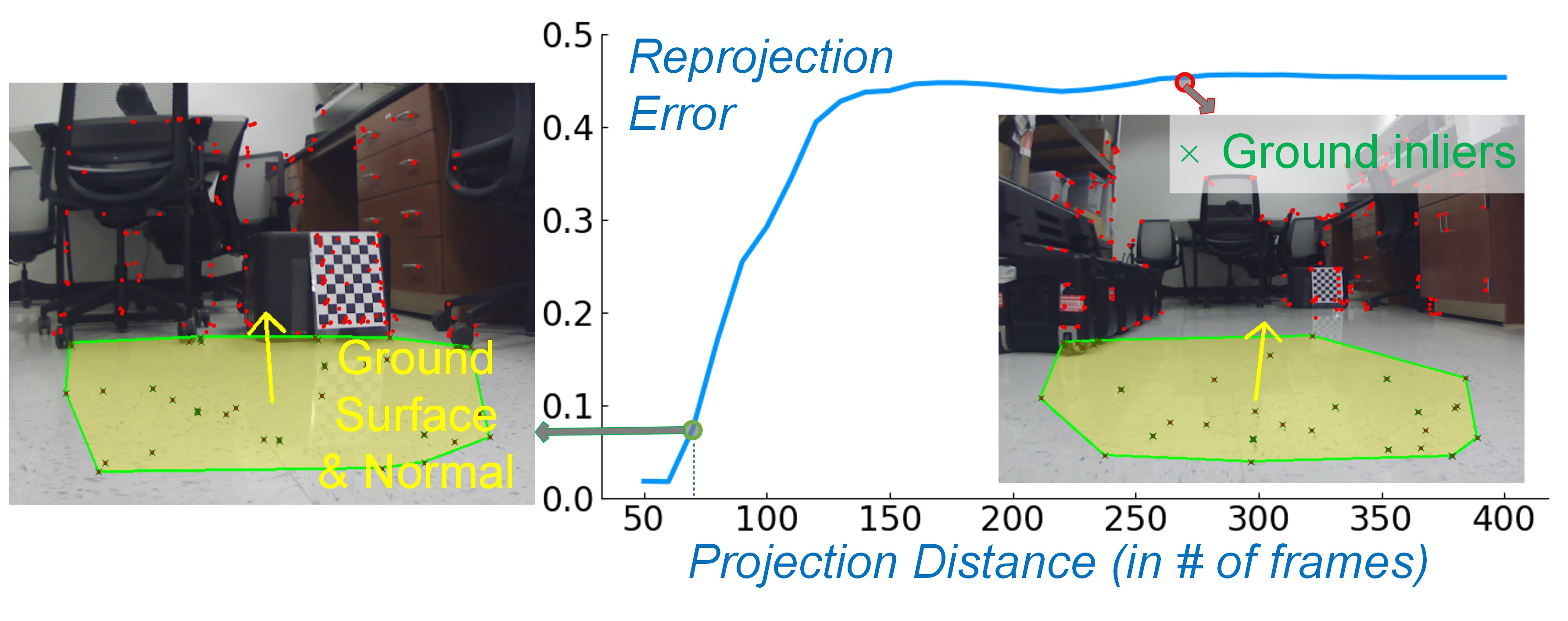}%
\vspace{-2.5mm}
\caption{\extension{Reprojection errors for reference points (checkerboard corners) are evaluated for different {EOB distances} ($f$). The estimated ground surface is shown as a convex hull of inlier points; the surface normal is overlaid for better visualization.}}
\label{fig:reprojection_b}
\end{subfigure}
\caption{We conduct 2D indoor navigation experiments with a TurtleBot4 to validate the geometric accuracy of our algorithm; here, results are visualized for ground plane estimation and reprojection errors of known reference points in the scene. %The improved EgoExo views for smaller \textit{EOB distances} are supported by other metrics such as lower better estimation of ground plane and lower checkerboard corner reprojection error.
}
\label{fig:reprojection}
\vspace{-2mm}
\end{figure}

\noindent
\textbf{Experimental setup}. The proof-of-concept experiments are conducted with TurtleBot4, a 2D ground robot that can be teleoperated with egocentric views from its front-facing monocular camera. It has only two degrees of freedom (DOF) for linear and angular velocity, which simplifies the motion kinematics for tracking its instantaneous position and orientation. We teleoperate it to collect visual data with a USB camera at $640\times480$\,p resolution in office, laboratory, and hallway scenarios. The experiments are designed to validate the proposed algorithm by evaluating ground plane estimation and reprojection errors. 

\vspace{1mm}
\noindent
\textbf{Geometric validation: reprojection error analysis}. We first evaluate the reprojection errors of known reference points from the generated EgoExo views and the estimated ROV pose. We use standard checkerboard corners as reference points from egocentric views and then evaluate the reprojection errors for those points from exocentric views. This test is iterated over different sets of past egocentric images, each corresponding to a different \textit{EOB distance}. As shown in Figure~\ref{fig:reprojection_a}, a checkerboard is viewed from different EOB distances (further back into the past), indicated by the parameter $f$. More specifically, $f$ is the number of frames between the current egocentric view and the selected EOB view. The corresponding reprojection error is plotted in Figure~\ref{fig:reprojection_b}, which shows how the estimation is accurate for lower values of $f$, and gradually degenerates for $f>100$. This is consistent with our visual observation of the projected ROV point cloud, \ie, it is on the ground plane with accurate orientation based on the SLAM trajectory estimates.

\begin{table}[t]
% \vspace{2mm}
\centering
\caption{\extension{Evaluation of ground plane estimation is presented for indoor UGV operation.}}
% \vspace{-1mm}
\footnotesize
\begin{tabular}{l||l|l|l|l}
  \Xhline{2\arrayrulewidth}
     \textbf{\# Ego} & \textbf{Inlier} & \textbf{Plane} & \textbf{Normal} & \textbf{Altitude} \\
     \textbf{Frames} & \textbf{Fraction} ($\uparrow$) & \textbf{RMSE} ($\downarrow$) & \textbf{Drift} ($\downarrow$) & \textbf{Drift} ($\downarrow$) \\
    \Xhline{2\arrayrulewidth} 
    $476$ & $99.4\%$ & $0.005$ & $4.67\degree$ & $0.058$ \\
    \Xhline{2\arrayrulewidth} 
\end{tabular}
\label{tab:indoor_eval}
\vspace{-2mm}
\end{table}

\vspace{1mm}
\noindent
\extension{\textbf{Geometric validation: ground plane estimation}. We adopt four metrics to evaluate the quality of ground plane estimation: inlier fraction, plane RMSE, temporal drift in plane normal, and temporal drift in altitude. The inlier fraction for each frame reports the ratio of inliers to total candidate points ($N$) after RANSAC fit:
\begin{equation}
\eta = \tfrac{1}{N}\sum_{j=1}^{N} \mathbf{1}\big(|\,\mathbf{n}^\top \mathbf{p}_j + d\,| < \tau \big),
\label{eqn:inlier_fraction}
\end{equation}
where $\mathbf{n},d$ are the fitted plane parameters, $\mathbf{p}_j$ are the candidate 3D points, and $\tau$ is the distance threshold (see Eqn.~\ref{eq:ransac_plane}). A higher $\eta$ indicates that the majority of candidate points are consistent with a single ground plane.
Subsequently, the point-to-plane distances of inliers are calculated to quantify the fitting residual as root-mean-square error (RMSE):
\begin{equation}
\text{RMSE} = \sqrt{\tfrac{1}{N}\sum_{j=1}^{N} \big(\mathbf{n}^\top \mathbf{p}_j + d \big)^2}.
\label{eqn:rmse}
\end{equation}
A lower RMSE reflects a tighter fit around the estimated plane.
Next, to assess temporal consistency, we compute the angular difference between consecutive plane normals:
\begin{equation}
\Delta\theta_i = \arccos\big( \tfrac{\mathbf{n}_i^\top \mathbf{n}_{i-1}}{\|\mathbf{n}_i\|\,\|\mathbf{n}_{i-1}\|} \big).
\label{eqn:normal_drift}
\end{equation}
The mean angular drift across frames is reported, where low values indicate temporal stability. Finally, the altitude at instance $i$ is computed as the vertical distance of the camera center $\mathbf{c}_i$ to the estimated plane:
\begin{equation}
h_i = \tfrac{\mathbf{n}_i^\top \mathbf{c}_i + d_i}{\|\mathbf{n}_i\|}.
\label{eqn:altitude_drift}
\end{equation}
In the absence of true measurement, the computed (scaled) altitude is not meaningful; however, a low deviation across frames indicates that the synthesized 2.5D ground remains consistent for visualization.}

\extension{In 2D indoor setup, the robot's camera is rigidly mounted with negligible roll and pitch variation, so the estimated ground-plane normal is expected to align with the camera's vertical axis and remain stable across frames. Consequently, the plane inlier fraction should be consistently high, and the residual error should approach zero. The results obtained from several trials in office, laboratory, and hallway scenarios are summarized in Table~\ref{tab:indoor_eval}. A high inlier fraction and low normal drift across frames confirm the robustness of our approach under such structured conditions; please refer to the next section for further evaluation in unstructured settings.} 

\extension{Figure~\ref{fig:reprojection_b} illustrates two representative examples from an office scene: one for $f=70$ with a low reprojection error, and another for $f=260$ with a high error. As seen, the estimated ground plane normal validates the geometric accuracy for the $f=70$ case. On the other hand, a misaligned plane normal for the $f=260$ case demonstrates the underlying error in pose estimation as well as in the reprojection process. Essentially, the geometric accuracy of our proposed algorithm depends on the pose estimation performance of the SLAM system.}

\begin{figure*}[h]
\centering
\begin{subfigure}[]{\linewidth}
\includegraphics[width=\linewidth]{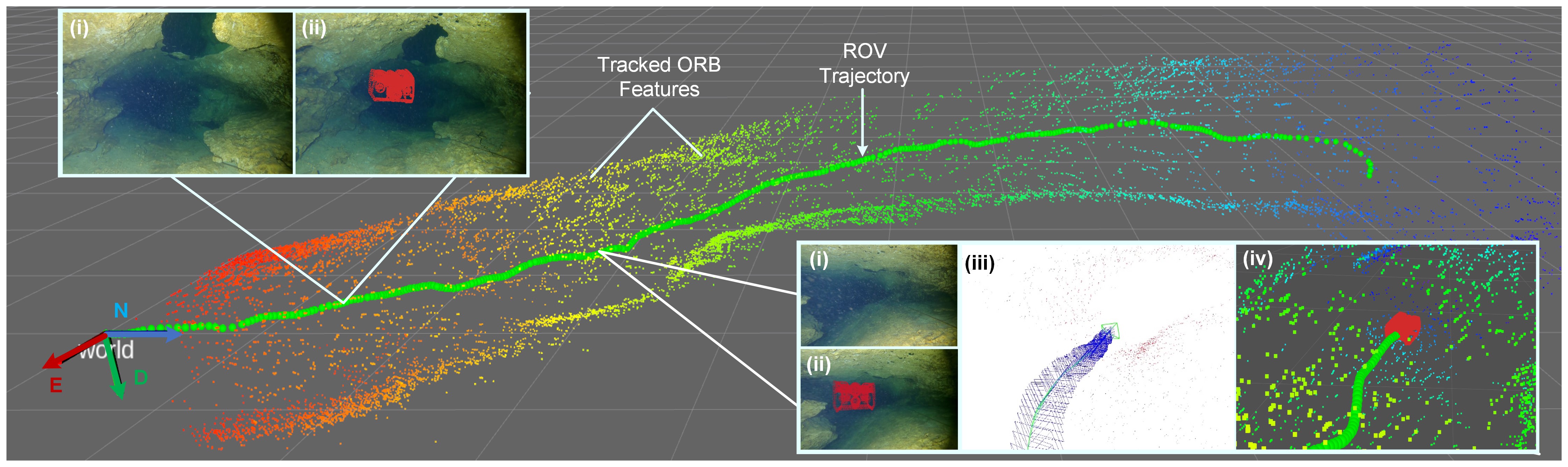}%
\vspace{-1.5mm}
\caption{EgoExo views: (i-ii) Ego and Exo views with rendered ROV pose; and (iii-iv) Updated camera poses and Exo view of the 3D map.}
\label{subfig:old_snap}
\vspace{1mm}
\end{subfigure}

\begin{subfigure}[]{\linewidth}
\includegraphics[width=\linewidth]{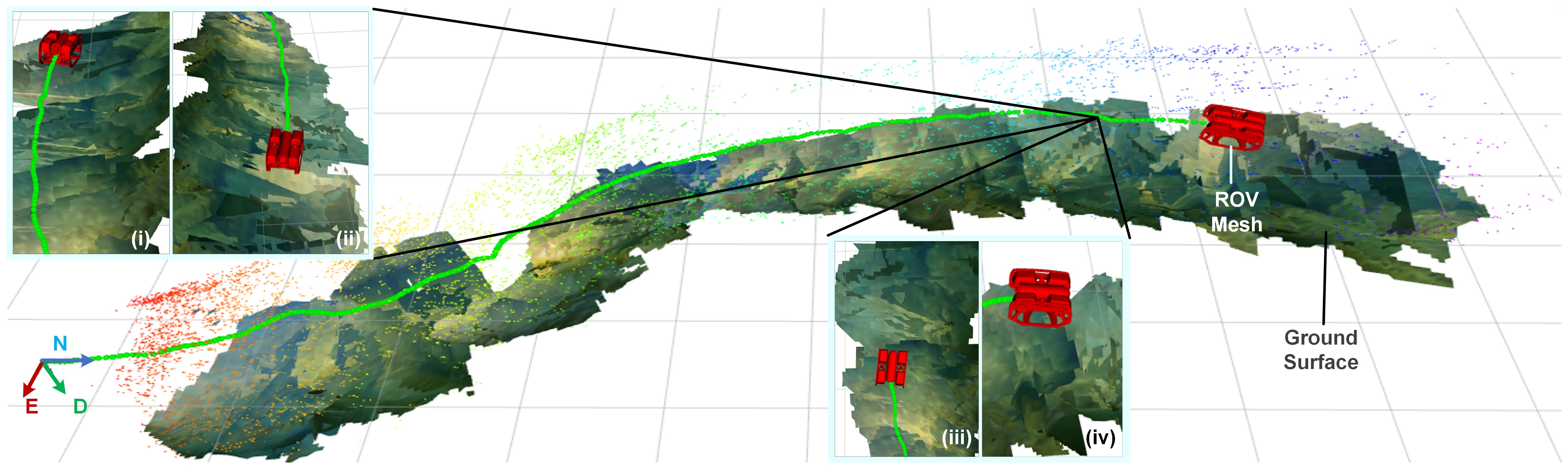}%
% \vspace{-1mm}
\caption{EgoExo++ views: operator-selected viewpoints above the ground surface- (i) back, (ii) front, (iii) top, and (iv) side. }
\label{subfig:new_snap}
\vspace{-1mm}
\end{subfigure}
\caption{\extension{EgoExo and EgoExo++ views are shown for field trials conducted in the Peacock Springs cave system, Florida. The EgoExo pipeline generates 2D exocentric imagery from directly behind the ROV, along with a sparse 3D map of the environment. The EgoExo++ extends it by reconstructing the ground surface and offering full $360\degree$ exocentric viewpoints.}}
\label{fig:snap}
\end{figure*} 

\begin{table}[h]
% \vspace{-1mm}
\centering
\caption{Memory requirement of the proposed framework for different buffer lengths.}
\footnotesize
\begin{tabular}{l||r|r|r|r|r}
  \Xhline{2\arrayrulewidth}
    \textbf{Buffer} (\# of Frames) & $50$ & $100$ & $200$ & $300$ & $400$\\  \hline
    \textbf{Memory Usage} (MB) & $65$ & $142$ & $301$ & $455$ & $609$ \\ 
    \Xhline{2\arrayrulewidth} 
\end{tabular}
\label{tab_memory}
\vspace{-4mm}
\end{table}

\vspace{1mm}
\noindent
\textbf{Computational efficiency}. We analyze the computational complexity of the proposed algorithm for different configurations to ensure real-time execution in resource-constrained edge devices onboard standard ROV platforms. Table~\ref{tab_memory} shows the memory requirement of our algorithm for different choices of buffer size. The memory footprint is \finalrev{$274$\,MB} for a buffer size of $180$ frames, making it highly efficient. \extension{Table~\ref{tab_map} demonstrates that the base EgoExo framework maintains a consistent output rate of over $25$\,FPS (frames per second). The added computation for ground estimation slightly reduces the scene update rate in EgoExo++, but still maintains over $20$\,FPS, making it suitable for integration in existing teleoperation engines. 
% In fact, our algorithm only adds $6.8\%$ overhead on the SLAM backbone for the global map update.
}  
%Additionally, the algorithm is executed  in real-time when integrated with a monocular SLAM backbone. We demonstrate this by maintaining a consistent output rate exceeding $20$\,FPS (see Table~\ref{tab_map}), which meets the requirements for almost all teleoperation applications.

\begin{table}[h]
% \vspace{2mm}
\centering
\caption{\extension{End-to-end computational speed of the proposed framework; the rows report: (\textbf{i}) ROS node publish rate of the imagery; and (\textbf{ii}) the global scene update rate.}}
\footnotesize
\begin{tabular}{l||c|c|c}
  \Xhline{2\arrayrulewidth}
    \textbf{Method} & \textbf{SLAM} & \textbf{SLAM+EgoExo} & \textbf{SLAM+EgoExo++} \\
    % & \textbf{Only} & \textbf{ } & \textbf{EgoExo} \\
     %&  &  & \textbf{EgoExo++} \\
     \Xhline{2\arrayrulewidth} 
    Image Update  & $26$\,FPS & $25.1$\,FPS & $25.1$\,FPS \\
    Scene Update & $26$\,FPS & $25.3$\,FPS & $20.2$\,FPS \\
    \Xhline{2\arrayrulewidth} 
\end{tabular}
\label{tab_map}
\vspace{-4mm}
\end{table}
% \begin{table}[h]
% \vspace{-1mm}
% \centering
% \caption{Integrating EgoExo on top of SLAM backbone adds to computational complexity. Here, the effect is quantified in terms of output publishing rate.}
% \begin{tabular}{c|cc|cc}
% \hline
% Process    & \multicolumn{2}{c|}{SLAM only}           & \multicolumn{2}{c}{SLAM + EgoExo}      \\ \Xhline{2\arrayrulewidth}     & \multicolumn{1}{c|}{Tracked image} & Map & \multicolumn{1}{c|}{EgoExo image} & Map  \\ 
%  O/P Rate & \multicolumn{1}{c|}{$26$ FPS}          & $26$ FPS  & \multicolumn{1}{c|}{$25.1$ FPS}              & $25.2$ FPS \\ \hline
% \end{tabular}
% \label{tab_2}
% \end{table}

\subsection{Field Deployment: 3D Underwater Caves\label{subsec:cave_exp}}
\noindent
\textbf{Experimental setup}.
We extend our experiments to underwater cave exploration scenarios, where the ROV performs \invis{possesses} full 6-DOF motions. While the $roll$ motion is limited in the standard BlueROV2, we consider all 6-DOF for teleoperation with the buoyancy change and pressure imbalance caused by water flow at the cave openings. For remote teleoperation, we consider the scenarios where human operators maneuver an underwater ROV from the surface by following the caveline and other navigation markers as guides~\cite{abdullah2023caveseg}. The mission objective is to navigate the ROV $75$-$300$ feet deep inside the cave through its complex structures, and then safely return it to the surface. The videos are recorded at $1920\times1080$\,p resolution with a GoPro11 camera mounted on BlueROV2 and then compressed to $640\times480$\,p within our framework. In addition to evaluating the geometric accuracy, we consider how informative the generated views are compared to traditional consoles for ROV teleoperation.   
%Although standard BlueROV2 robots are not capable of $roll$, practical observations  reveal frequent rolling motion resulting from factors such as weight imbalance, buoyancy change, or pressure imbalance caused by water flow at the cave openings etc. Hence we take into account all six degrees of motion for evaluation. 

% a sample image pair where the homography estimation is visualized with a 2D square logo projected onto a reference plane in EgoExo view. All the corners cannot be detected properly in such noisy low-light conditions. However, the projected logo shows a fair estimation of homography even with the partial detections. The accuracy of estimated homography from Ego-Exo pair further indicates the underlying geometric validity of our ROV projection method to generate exocentric views.

%between a current frame $c$ and a reference frame $r$. The homography matrix $H$ is given by,

% \begin{equation}
%     \mathbf{H} = \mathbf{R}\, - \mathbf{t}\cdot \mathbf{n}^{T} 
% \end{equation}

% Where ,
% \begin{equation}
%     \mathbf{H} = ({}^{w}_{r}\mathbf{R}^{-1}\, {}^{w}_{c}\mathbf{R})\cdot \mathbf{P} + ({}^{w}_{c}\mathbf{t} - {}^{w}_{r}\mathbf{t}),
% \end{equation}

\vspace{1mm}
\noindent
\textbf{Real-time map update and teleoperation}.
In addition to the exocentric view generation and ROV pose rendering, the EgoExo framework simultaneously updates a sparse map with extracted feature points from the SLAM system. Figure~\ref{fig:snap} shows an ROV's trajectory mapped during our trial in an underwater cave in Peacock Springs, Florida. As seen, the generated EgoExo views embed more peripheral information about the scene. The exocentric view of the ROV pose and its relative distance from cave walls or overhead obstacles are useful to surface operators for obstacle avoidance and efficient decision-making. Additionally, the 3D map shows the ROV's past trajectory and its current pose, which are useful to analyze the mission progress, which is not possible in traditional teleoperation consoles. Such a global view of the trajectory is also useful during emergency evacuation and recovery. Beyond cave exploration, these features will be crucial in ROV-based subsea surveillance and search-and-rescue operations as well.

\begin{figure}[t]
\centering
%\vspace{-2mm}
\includegraphics[width=\linewidth]{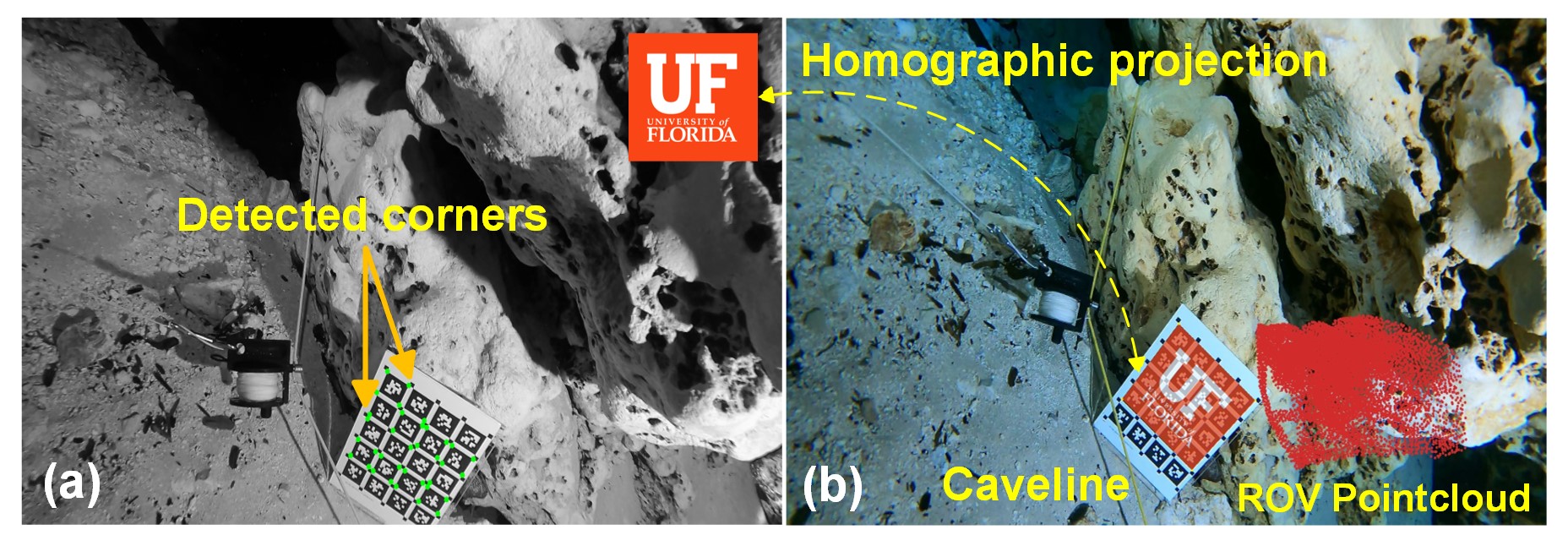}%
\vspace{-3mm}
\caption{A snapshot from our cave exploration scenario: (a) Egocentric view with detected reference points; and (b) Synthesized EgoExo view with projected ROV point cloud. We use a sample logo for homographic projection on the reference surface to demonstrate the accuracy in pose estimation. 
%and Reference corner points (marked in green) are detected from robot's egocentric view. (b) The estimated homography in EgoExo view is visualized with a logo projection. Note that, some corners are not visible in egocentric view.
}
\label{fig:homography}
\vspace{-2mm}
\end{figure}

%in (b) have clear advantages over their corresponding egocentric images in (a). The EgoExo views show the ROV's relative position and distance from cave walls and overhead obstacles, making it easier for the operator to avoid collision and safely navigate through challenging cave segments. Additionally, the 3D map shows the ROV's past trajectory and its current location from a global point of view which are useful to analyze the mission's progress and current status, compared to the single egocentric view-based system where the operator often remains clueless about the ROV's actual location. During emergency, this existing trajectory map also serves as an additional marker that the operator can follow to safely exit the cave.
%The interactive map also allows the operator to zoom into particular segments and keep note of any markers or obstacles found. 
%The also  to follow when exiting the cave. 

\vspace{1mm}
\noindent
\textbf{Validation: homographic projection}. Due to the complex scene geometry inside underwater caves, we adopt a homography estimation approach for the performance validation. As shown in Figure~\ref{fig:homography}, April-Tag~\cite{olson2011apriltag} corners are used as reference points for reprojection. Specifically, we compute the homography transformation between the egocentric and synthesized exocentric views to visualize the reprojection errors. We use a sample 2D logo and project it onto the reference April-Tag surface using the homographic transform. The unskewed planar projection validates the accuracy of the pose estimation and point cloud rendering processes.

\vspace{1mm}
\noindent
\extension{\textbf{Validation: ground plane and 2.5D view}. The ground plane estimation in the field is assessed following the same method as the indoor validation; Table~\ref{tab:ground_plane_eval} summarizes the results for trials performed at two different cave systems. In addition to the four metrics defined earlier in Eqn.~\ref{eqn:inlier_fraction}-\,\ref{eqn:altitude_drift}, a success rate is also reported, since a valid ground plane cannot always be recovered in unstructured cave scenes. The success rate is defined as the percentage of frames in which a plane can be reliably fitted from the sparse SLAM feature points.
}

\begin{table}[h]
% \vspace{2mm}
\centering
\caption{\extension{Evaluation of ground plane estimation is presented for field trials conducted in two distinct cave systems in FL, USA.} }
\footnotesize
\begin{tabular}{l||c|c}
  \Xhline{2\arrayrulewidth}
    \textbf{Field Trials} & \textbf{Peacock Springs} & \textbf{Devil's Springs } \\\Xhline{2\arrayrulewidth} 
    Ego Frames   & $485$ segments  & $1153$ segments \\ \hline
    Success Rate ($\uparrow$)  & $85.8\%$ & $97.17\%$ \\
    Inlier Fraction ($\uparrow$)  & $94.0\%$  & $90.7\%$  \\
    Plane RMSE ($\downarrow$) & $0.04$  & $0.08$ \\
    Normal Drift ($\downarrow$) & $6.3\degree$ & $20.49\degree$ \\
    Altitude Drift ($\downarrow$) & $0.29$ & $0.47$ \\
    \Xhline{2\arrayrulewidth} 
\end{tabular}
\label{tab:ground_plane_eval}
\vspace{-2mm}
\end{table}

\extension{
The results in Table~\ref{tab:ground_plane_eval} show that although trials in Devil's Springs cave systems have a higher success rate in detecting the ground plane, the plane quality from Peacock Springs cave systems is consistently better. This difference can be attributed to the more complex cave structure and the challenging ROV trajectory executed in the latter case. In Devil's Springs, several obstacles (\eg, large rocks) appeared directly in front of the ROV, forcing the operator to ascend and maneuver around. The terrain itself had high altitude variations, composed of rocks, boulders, and scattered pebbles, in contrast to the relatively smooth sedimentary floor observed in Peacock Springs. The reconstructed ground map from Peacock Springs shows consistent elevation and orientation (see Figure~\ref{subfig:new_snap}), supporting the quantitative results. More snapshots from the two sites are provided in Figure~\ref{fig:viewpoint}. As seen, the rocky terrain in Devil's Springs and the resulting jerky vehicle motion led to larger errors in ground plane estimation, greater deviations in the fitted normal, and higher variability in estimated altitude.
}

% \begin{figure*}[t]
% \centering
% \includegraphics[width=\linewidth]{imgs/bev_snap.jpeg}%
% \vspace{-1mm}
% \caption{The reconstructed ground skin }
% \label{fig:bev_snap}
% \vspace{-3mm}
% \end{figure*} 

\vspace{1mm}
\noindent
\textbf{Observations: strengths and limitations}. Our experiments reveal some key strengths of the proposed teleoperation framework. First, the generated exocentric views closely resemble the actual EOB views during a smooth trajectory, which is usually the case for subsea exploration and surveying tasks. Second, the buffer memory works as a backup during a temporary failure of the SLAM system, typically observed at turning corners or due to abrupt motion. \finalrev{In such cases, our algorithm retains historical poses along with their associated egocentric images from its buffer memory. Unlike the sparse SLAM map and last known poses, which are geometrically sparse and semantically empty, the buffered images and the 2.5D terrain map provide richer spatial context, assisting the operator to understand the recent scene layout better} and safely anchor or pause the mission until communication is restored. On the other hand, its heavy dependency on the SLAM backbone leads to some inherent limitations. Feature-based monocular SLAM systems often fail in feature-deprived, noisy underwater scenes, which leads to inaccurate pose tracking and thus inaccurate EgoExo view synthesis. Tracking 6-DOF ROV motion from monocular vision is particularly challenging with no additional sensor to recover the scale information~\cite{JoshiIROS2019,wu2023sdu_sfm}. We observe some instances where the estimated ROV pose is incorrectly scaled in the rendering. To address this, multi-sensor fusion-based underwater SLAM backbones~\cite{RahmanIJRR2022,mo2021fast,lago2024visual} can be utilized in more critical applications.

\begin{table*}[t]
% \vspace{-2mm}
    \centering
    \caption{In our study, $15$ human participants provide their feedback to the following two sets of questions: (\textbf{i}) The first $10$ questions are from SUS~\cite{brooke1996sus}; and (\textbf{ii}) The remaining three questions are custom-designed. Response to each question is scaled from $1$ (strongly disagree) to $5$ (strongly agree).}
    \footnotesize
    \begin{tabular}{
        >{\centering\arraybackslash}p{0.05\textwidth}
        >{\arraybackslash}p{0.7\textwidth}
        >{\raggedleft\arraybackslash}p{0.15\textwidth}}
        \toprule
        \textbf{\#} & \textbf{Questions} & \textbf{Mean,~Std. Dev.} \\
        \midrule
        %\multicolumn{3}{c}{\textbf{System Usability Scale (SUS)}~\cite{brooke1996sus}} \\\midrule
        1 & I think that I would like to use this system frequently. & 4.3$,~$0.6 \\
        2 & I found the system unnecessarily complex. & 2.0$,~$0.7 \\
        3 & I thought the system was easy to use. & 4.3$,~$0.4 \\
        4 & I think that I would need the support of a technical person to be able to use this system. & 2.0$,~$0.6 \\
        5 & I found the various functions in this system were well integrated. & 4.0$,~$0.8 \\
        6 & I thought there was too much inconsistency in this system. & 2.3$,~$0.7 \\
        7 & I would imagine that most people would learn to use this system very quickly. & 4.4$,~$0.5 \\
        8 & I found the system very cumbersome to use. & 2.0$,~$0.6 \\
        9 & I felt very confident using the system. & 3.7$,~$0.7 \\
        10 & I needed to learn a lot of things before I could get going with this system. & 1.4$,~$0.5 \\
        \midrule
        %  \multicolumn{3}{c}{\textbf{Custom Questions}} \\
        % \midrule
        11 & The proposed exocentric view is beneficial for ROV teleoperation. & 4.5$,~$0.5 \\
        12 & I found the EOB distance tuning feature useful to get the best view. & 4.5$,~$0.5 \\
        13 & The generated 3D map provides a better understanding of the ROV's global location and its surroundings. & 4.6$,~$0.9 \\
        \bottomrule
    \end{tabular}
    \label{tab:study}
    \vspace{-3mm}
\end{table*}

\begin{table*}[t]
% \vspace{-2mm}
    \centering
    \caption{\rebuttal{NASA-TLX subjective workload is compared between egocentric-only teleoperation and EgoExo++. Scores are collected from $15$ participants on the standard unweighted $0$ (very low) to $20$ (very high) scale; \textbf{Mean, Standard Deviation} are reported.}}
    \footnotesize
    \begin{tabular}{
        >{\centering\arraybackslash}p{0.05\textwidth}
        >{\arraybackslash}p{0.47\textwidth}
        >{\raggedleft\arraybackslash}p{0.16\textwidth}
        >{\raggedleft\arraybackslash}p{0.16\textwidth}
        }
        \toprule
        \textbf{\#} & \textbf{Questions} & Ego-only (Baseline) & \textbf{EgoExo++ (Proposed)}\\
        \midrule
        1 & How mentally demanding was the task? & $9.3,\,4.9$ & $6.1,\,3.5$ \\
        2 & How physically demanding was the task? & $6.2,\,5.8$ & $4.1,\,3.7$ \\
        3 & How hurried or rushed was the pace of the task? & $8.1,\,4.8$ & $7.3,\,5.5$ \\
        4 & How successful were you in accomplishing what you were asked to do? & $14.3,\,3.2$ & $17.0,\,2.5$ \\
        5 & How hard did you have to work to accomplish your level of performance? & $9.2,\,4.4$ & $5.4,\,3.1$ \\
        6 & How insecure, discouraged, irritated, stressed, and annoyed were you? & $3.9,\,2.8$ & $2.3,\,2.1$ \\
        % \hline
        %   & Overall    $8.5\pm4.3$   & $7.03\pm3.4$

        % \midrule
        %  \multicolumn{3}{c}{\textbf{Custom Questions}} \\
        % \midrule
        % 7 & & & \\
        % 8 &  & & \\
        % 9 &  & & \\
        \bottomrule
    \end{tabular}
    \label{tab:nasa_tlx}
    \vspace{-1mm}
\end{table*}

\subsection{User Study \#1: SUS}
\finalrev{The goal of this user study is to assess interface-level usability using real mission data. Hence, the} study is conducted with multiple underwater cave exploration data collected during our field trials. A BlueROV2 recorded egocentric video feeds inside the caves at up to $100$-meter penetrations. 
% Later on, $15$ human participants, between the ages of $21$-$32$ with little/no prior teleoperation experiences, evaluate the ease of operation with our developed console and compare it to traditional consoles.
\finalrev{Later on, the playback sessions are presented to $15$ human participants, between the ages of $21$-$32$, with little/no prior teleoperation experiences. They evaluate the ease of operation with our developed EgoExo console and compare it to traditional consoles.}
Their feedback is recorded using the System Usability Scale (SUS)~\cite{brooke1996sus}, with our interface achieving an average SUS score of $77.5$. We also formulate an independent set of questions on the teleoperator's preference for the novel features of our method. The individual questions and corresponding scores are presented in Table~\ref{tab:study}. Some key observations from this study are listed below.

\begin{enumerate} [label={\arabic*)},nolistsep,leftmargin=*] 
\item The obtained SUS score is fairly above median (score:~$68$) and is considered {\tt Good} for user experience; it is slightly below the {\tt Excellent} (score:~$80.3$) category.

\item Post-operation feedback from our ROV operators suggests that the exocentric views are more useful for safe ROV maneuvers. 

\item The synthesized 3D map provides a better sense of the ROV's global location and improves spatial awareness of the teleoperators.

\item The operators report a significantly lower workload (perceived cognitive load) in conducting complex tasks such as object following and structure mapping. 
\end{enumerate}

\begin{figure}[t]
    \centering
    \includegraphics[width=\linewidth]{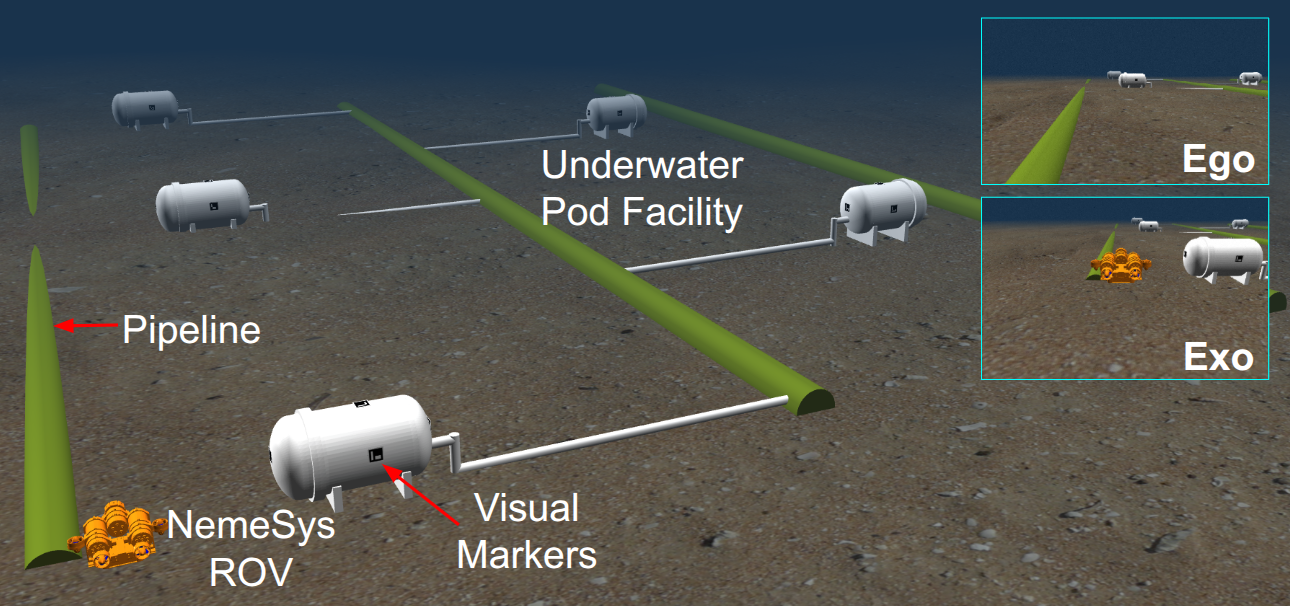}
    \vspace{-1mm}
    \caption{\rebuttal{A Gazebo-simulated underwater facility with five subsea pods is used to evaluate teleoperation performance. Participants drive the ROV by following the green pipelines and visually identify fiducial markers mounted on the pods. Each operator repeats the mission twice -- first using the egocentric camera feed and then using the proposed EgoExo++ interface.}}
    \vspace{-1mm}
    \label{fig:sim_world}
\end{figure}

\subsection{User Study \#2: NASA-TLX}
\vspace{-0.5mm}

\finalrev{Unlike the SUS study, this experiment involves active teleoperation, enabling objective assessment of performance metrics such as mission completion time, path deviation ratio, and collision count. Such metrics are difficult to measure in real cave environments due to the lack of reliable ground-truth localization and trajectory data. Hence,} this user study is conducted in a Gazebo simulated environment of a $60$\,m deep, $25$\,m$\times$$15$\,m underwater industrial facility containing pipelines, subsea pods~\cite{abdullah2025active,blow2025detection}, and visual fiducial markers (see Figure~\ref{fig:sim_world}). Participants are tasked with teleoperating the NemeSys robot in ROV mode~\cite{abdullah2025nemesys} in a \textit{lawnmower} pattern along the three green pipelines and visually detecting the front/back-faced markers attached to the pod with the ROV's front-facing camera. The autopilot assists in maintaining the intended depth, so the operator primarily controls surge and yaw motion in the horizontal plane; however, depth and roll control remain available for collision avoidance if needed. The participant pool consists of $15$ individuals (aged $25$-$36$), including $4$ females and $11$ males, with prior teleoperation experience distributed as $3$ experts (significant ROV teleoperation experience), $4$ intermediate users, and $8$ novices (no teleoperation experience).
Each participant performs the same mission twice: first using only egocentric camera views and then using the proposed EgoExo++ system.

\vspace{1mm}
\noindent
\rebuttal{\textbf{Subjective evaluation}. Subjective workload is assessed using NASA-TLX~\cite{hart1988development}, collected on an unweighted $0$ (very low) -- $20$ (very high) scale; the results are summarized in Table~\ref{tab:nasa_tlx}. EgoExo++ reduces perceived workload across all six dimensions. For instance, participants report $34\%$ reduction in mental demand and feel $19\%$ more successful when using our teleoperation console. These results suggest that EgoExo++ reduces cognitive effort and stress while improving task efficiency.
}

\begin{table}[h]
    \centering
    \caption{\rebuttal{Task performance for $15$ users is compared between the traditional egocentric teleoperation interface and EgoExo++. Scores for mean and standard deviation are reported for the first two metrics, while the final row reports the total collision count across all participants.}}
    \footnotesize
    \begin{tabular}{c||c|c}
        \Xhline{2\arrayrulewidth}
         Metric & Ego-only & \textbf{EgoExo++} \\
         & (Baseline) & \textbf{(Proposed)} \\
         \Xhline{2\arrayrulewidth}
         Average mission time (Sec. $\downarrow$)  & $145.3\pm31$ & $121.3\pm20$ \\
         % Path efficiency (\%) ($\uparrow$) & $89.3\pm0.05$ & $95.2\pm0.06$ \\
         PDR: Path Deviation Ratio (\% $\downarrow$) & $12.3\pm6.0$ & $2.5\pm9.3$ \\
         Total collision count ($\downarrow$)  & $5$ & $2$ \\
         \Xhline{\arrayrulewidth}
    \end{tabular}
    \label{tab:teleop_metrics}
    \vspace{-3mm}
\end{table}

\vspace{1mm}
\noindent
\rebuttal{\textbf{Objective evaluation}. We report three objective metrics to assess task performance; the results are summarized in Table~\ref{tab:teleop_metrics}. Mission time measures the duration required to navigate past all $5$ pods and detect all $10$ target markers. 
% Path efficiency is defined as the ratio between the ideal path length and actual driven length, expressed as a percentage; higher values indicate tighter adherence to the intended lawnmower trajectory.
Path Deviation Ratio (PDR) represents the normalized overhead distance beyond the ideal path length ($81.75$\,m); it is calculated as: $\text{PDR} = (L_{actual} - L_{ideal}) / L_{ideal} \times 100\%$.} 

\rebuttal{
Essentially, $\text{PDR} = 0$ indicates perfect adherence to the intended trajectory, while positive values represent proportionally higher deviation. Collision is counted as the number of physical contacts with the seabed, pipelines, or pod structures. Compared to egocentric-only teleoperation, EgoExo++ achieves $16\%$ faster mission completion and $5\times$ lower PDR. \finalrev{The relatively large standard deviation in PDR reflects variability in operator expertise, particularly among novice participants who made wide turns at the sharp corners.} \finalrev{Additionally, only $2$ collisions were observed under EgoExo++ compared to $5$ under egocentric-only teleoperation,} indicating safer and more confident navigation behavior of the ROV.
}

\vspace{1mm}
\noindent
\finalrev{\textbf{Order bias and counterbalanced validation}. To reduce potential order bias and learning effects between two consecutive trials, we use a practice/familiarization phase before the formal experiments. Specifically, we brief the participants about the facility layout, mission objective, and target locations. They are then allowed to freely teleoperate the ROV using any of the available visualization modes (Ego-only, trailing EgoExo, and ground-referenced map). This practice session allows participants to become comfortable with the joystick controls and the visual interfaces before data collection begins. Furthermore, we conduct a supplementary evaluation in which $5$ participants ($2$ new and $3$ from the previous pool) perform the trials in reversed order (EgoExo++ first, then Ego-only). The results show consistent trends with the main study: EgoExo++ again achieves faster mission completion (approximately $18\%$ improvement) and lower path deviation ratio ($2.6\%$) compared to the Ego-only baseline ($9.0\%$). These results suggest that the observed performance gains are independent of the order in which the trials are performed.}

% \JI{``over-travel" sounds weird. Trajectory Overhead (TO) or Path Redundancy (RI) might be better. Also the Path Deviation Ratio (PDR) would probably sound better than path efficiency. PDR = (actual-ideal)/actual}

% \JI{citep was not used, I changed the cite to citep - make sure other sections have right use of cite/citep}

% \Adnan{My bad, the terms were misleading. Your formula of PDR is the same as what I defined as over-travel. I have changed the name to PDR. I removed the other efficiency metric since it showed only a $7\%$ improvement - sounds insignificant.}

\begin{figure}[t]
% \vspace{-3mm}
    \centering
    \includegraphics[width=\linewidth]{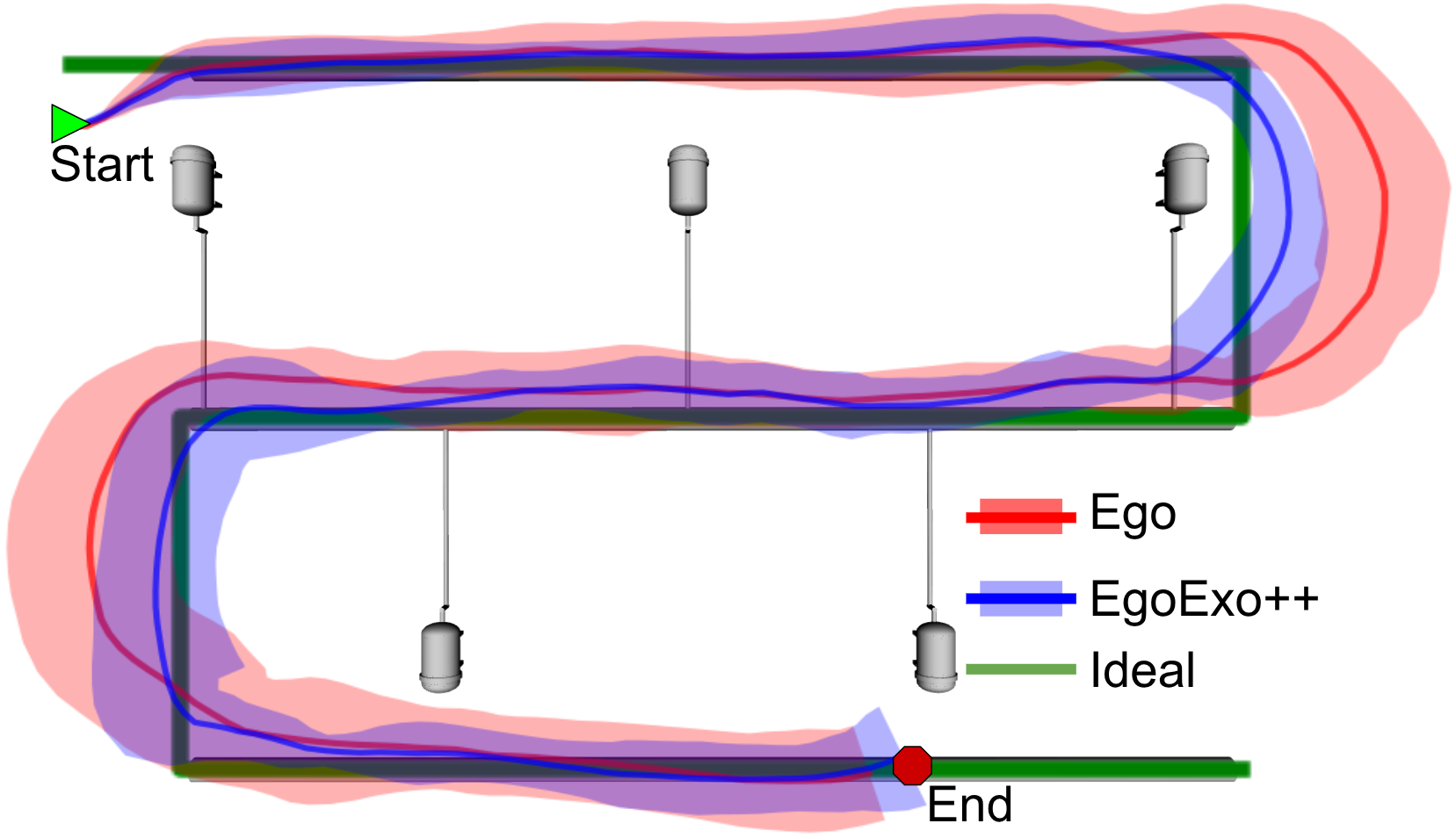}
    \vspace{-2mm}
    \caption{\rebuttal{ROV trajectories for egocentric and EgoExo++ teleoperation are visually compared. Egocentric-only missions (red) exhibit wider turning radii and more lateral deviation (shaded red regions). In contrast, EgoExo++ trajectories (blue) are more compact and aligned with the ideal pipeline route.}}
    \label{fig:sim_result}
    \vspace{-2mm}
\end{figure}

\vspace{1mm}
\noindent
\rebuttal{\textbf{Qualitative insights}. The aggregated trajectory overlays in Figure~\ref{fig:sim_result} further support the quantitative findings: EgoExo++ trajectories are more compact and aligned with the intended route, whereas ego-only trajectories exhibit wider turns and more lateral deviations. Moreover, qualitative feedback from participants provides two key insights: 
\begin{enumerate} [label={(\arabic*)},nolistsep,leftmargin=*] 
\item The EgoExo trailing view helps operators anticipate turning points and align the ROV efficiently to identify pod markers with that peripheral information; and
\item The 2.5D EgoExo++ map offers a global situational context, allowing operators to maintain a clear sense of ``where am I?'', particularly during U-turns where no pipeline is visible.
\end{enumerate}
Nevertheless, a commonly reported limitation of our system is perceptual disorientation caused by occasional latency in the augmented view. Aside from this effect, the operators preferred EgoExo++ for more confident ROV teleoperation.
}

% In addition to subjective workload assessment via NASA-TLX~\cite{hart1988development} (see Table~\ref{tab:nasa_tlx}), we record three objective metrics: (i) mission completion time, (ii) path efficiency, and (iii) collision frequency (see Table~\ref{tab:teleop_metrics}). Across participants, EgoExo++ leads to a measurable reduction in perceived workload, while operators demonstrate x\% faster mission completion, x\% efficient trajectories, and x\% fewer collisions compared to egocentric-only teleoperation. The aggregated trajectories shown in Fig.~\ref{fig:sim_result} also support the quantitative analyses: the EgoExo++ trajectories are more compact and consistent with the intended route. 

\section{Improved Underwater ROV Teleoperation: Strengths, Challenges, and Limitations}
\begin{figure*}[t]
    \centering
    \includegraphics[width=0.98\linewidth]{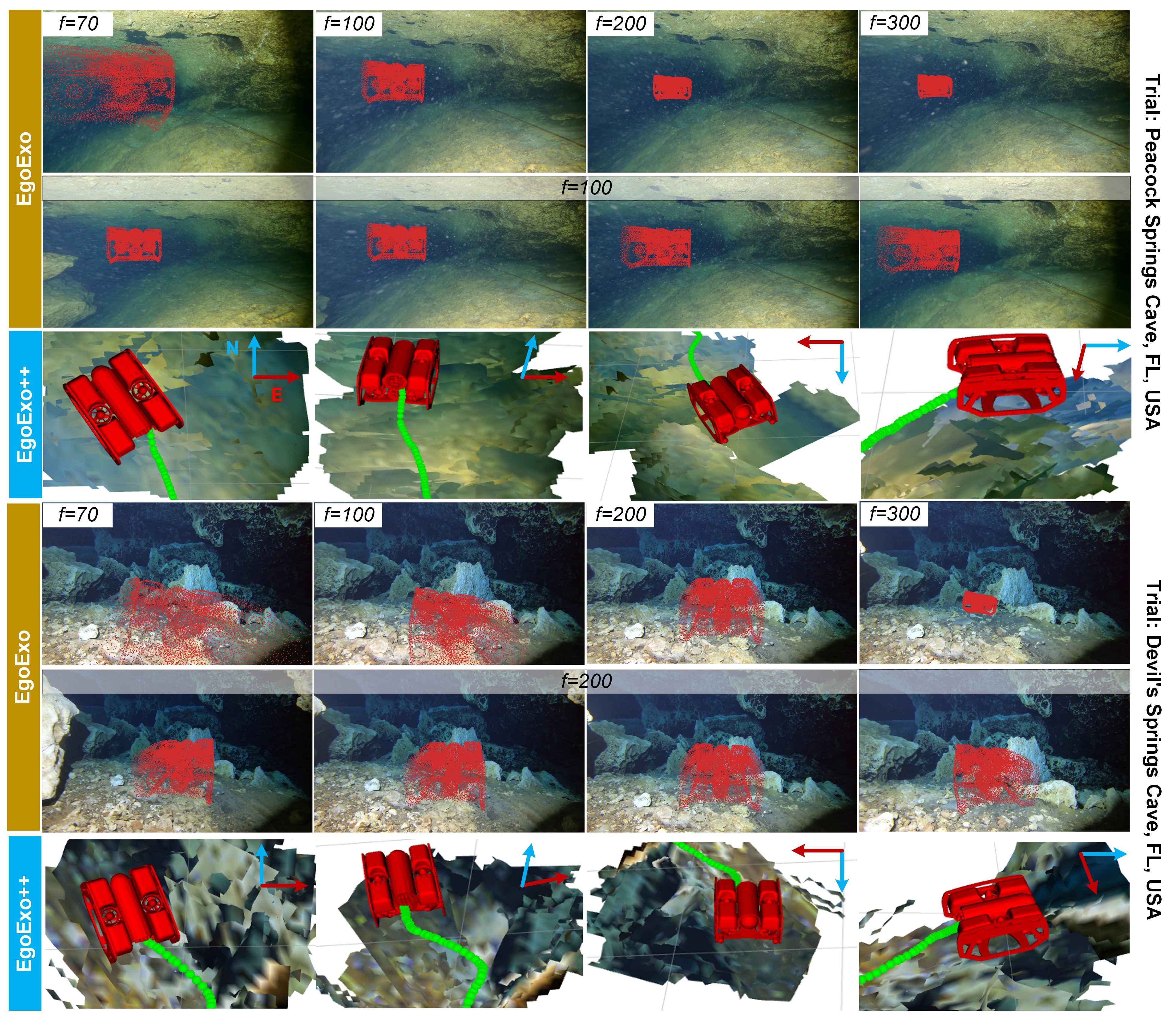}%
    \vspace{-2.5mm}
    \caption{\extension{EgoExo and EgoExo++ views are shown from field trials at two different cave systems. In EgoExo, the operator slides across the \textit{EOB distance $f$} to find the preferred Exo view, \eg, $f=100$ for the first case. EgoExo++ further enables free $360\degree$ viewpoint control, allowing the ROV to be visualized from arbitrary perspectives such as top, back, front, and side views. A video demonstration can be seen here: \url{https://youtu.be/xpvnzIJ_YbM}.}}
    \vspace{-2mm}
    \label{fig:viewpoint}
\end{figure*}

\noindent
\textbf{Multiple augmented viewpoints}.  We validate the utility of our proposed teleoperation interface through further experiments on underwater cave exploration data. Our expedition in cave segments at Devil's Springs, Florida, reveals that when ROVs move slowly against strong currents, extending the exocentric viewpoint distance can significantly improve teleoperation. This is achieved by tuning the queue parameters $r$, $c$, and $n$ in the proposed teleoperation interface. We consistently find that exocentric views are more informative, especially for about $5$-$10$ seconds preceding the ROV position during navigation. The multiple preceding views offered by our interface are particularly useful for mapping large structures such as newly discovered cave segments or shipwrecks~\cite{eustice2005large,ChatzispyrouICRA2026,chatzispyrou2025mapping}. As Figure~\ref{fig:viewpoint} shows, the synthesized viewpoints provide more spatial context, enabling operators to control the ROV efficiently around complex underwater structures.  

\vspace{1mm}
\noindent
\extension{\textbf{2D and 2.5D exocentric view}.
Our EgoExo pipeline synthesizes third-person views as flat 2D projections of the robot model onto a reference egocentric view from the past. The EgoExo++ advances from this purely image-based rendering to a 2.5D representation by recovering and texturing a dense ground surface in 3D space. This addition provides both altitude awareness and geometric context relative to the terrain, enabling operators to maintain safe clearance above uneven ground~\cite{hing2010chaseview}. Insights from our prior user study also emphasized the value of adjustable viewpoints, such as bird's-eye or side perspectives. While earlier EOB viewpoints offered multiple exocentric views, they remained anchored to the robot's trajectory and fixed reference frames. The EgoExo++ view is no longer tied to a fixed reference image; the virtual viewpoint can be freely adjusted, offering better situational awareness during ROV teleoperation~\cite{lager2018remote}.
% In this sense, EgoExo++ completes the exocentric view synthesis pipeline by producing a richer, geometry-aware visualization that is directly useful for ROV teleoperation.
}

% \begin{figure}[t]
% \centering
% \includegraphics[width=\linewidth]{imgs/viewpoint.jpeg}%
% % \vspace{-1mm}
% \caption{A demonstration of our \textit{adjustable EOB viewpoint} feature is shown. The $f$ numbers indicate the \textit{EOB distance} from current to the reference frame used for the Ego-to-Exo view generation. 
% Teleoperators can \textit{slide} across the distance ($f$) and find the best exocentric view, which is $f=200$ in this example.
% %$200$ looks near perfect, and anything beyond $300$ is barely useful.
% %leading to view occlusion by the ROV. The ROV's structure and pose looks much better from $200$ preceding frames. It looks too small from more than $250$ frames behind.
% }
% \label{fig:viewpoint}
% \vspace{-2mm}
% \end{figure}

\begin{figure}[t]
\centering
\includegraphics[width=\linewidth]{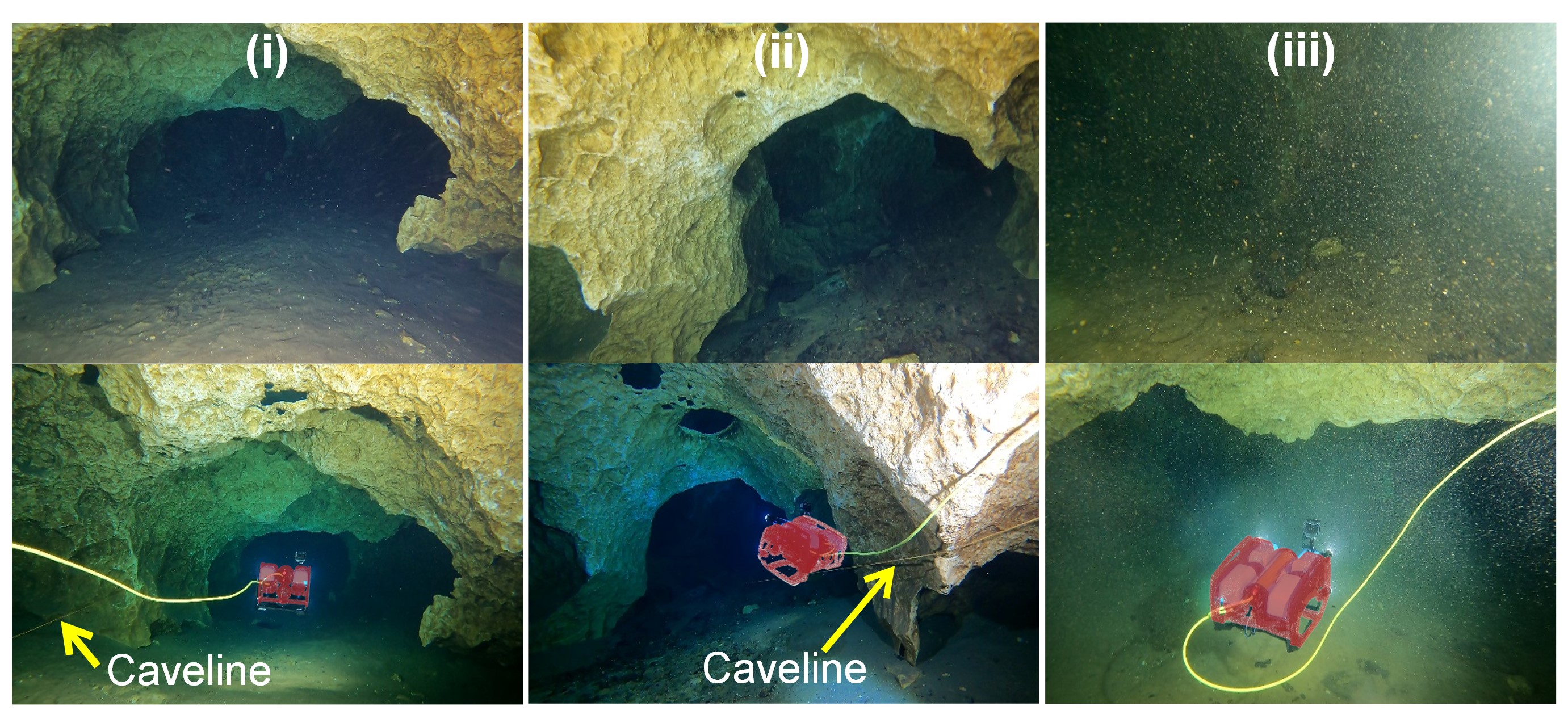}
% \vspace{-2mm}
\caption{Three challenging scenarios are shown for ROV teleoperation inside underwater caves: (\textbf{i}) caveline is not visible, \ie, blended with the background; (\textbf{ii}) caveline is not in the FOV; and (\textbf{iii}) front camera-light interactions with suspended particles are causing hazy egocentric views. In all cases, our augmented visuals are clearer and more informative to a surface operator.
}
\label{fig:utility}
\vspace{-3mm}
\end{figure}

\vspace{1mm}
\noindent
\textbf{Efficient teleoperation in complex missions}. We conducted extensive field trials across multiple underwater cave systems, including Orange Grove, Devil's Springs, and Peacock Springs, as well as inside a grotto system in Hudson, Florida.\invis{, accompanied by two human divers who recorded reference EOB views with Go-Pros.} 
% When only egocentric views were available, the operators reported that maneuvering the robot by following the caveline was quite challenging  because little/no ambient light penetrates inside underwater caves. 
We observe that maneuvering the robot by following the caveline with egocentric views is challenging because little/no ambient light penetrates inside underwater caves.
Despite using powerful lights, problems such as moving shadows and scattered waves create significant blind spots~\cite{gupta2025demonstrating}. Consequently, tracking and following the caveline or any other navigation markers~\cite{abdullah2023caveseg} without any peripheral view is extremely disorienting to the operator. In some cases, we observe that the cavelines get blended with the texture and features of cave walls in noisy conditions; see Figure~\ref{fig:utility}. In such scenarios, shifting the viewpoint to exocentric views allows easier identification of cavelines against the surrounding and overhead cave walls. Additionally, the augmented 3D map displays the robot's pose, allowing much safer maneuvering of the vehicle to its desired orientation~\cite{stewart2016interactive}.

\vspace{1mm}
\noindent
\textbf{Safer navigation in hazy low-light conditions}. Underwater caves present a unique formation of silt and sediment on their floor that results from erosion over extended periods. The silt is susceptible to disturbance from external factors~\cite{massone2024novel}, such as the motion of underwater ROVs or the turbulence generated by their propellers. Although ROV operators pay close attention to avoid contact with the floor and cave walls, it is often unavoidable due to buoyancy imbalance and strong flow of water. Dislodging the sediments results in cloudy or hazy conditions that obscure visibility~\cite{yu2022udepth}. Bright lights from the ROV reflect from these suspended particles and make it even more challenging to capture clear imagery of the surroundings. In such cases, third-person EOB views from behind the ROV offer a clearer and more informative perspective for navigation, as shown in Figure~\ref{fig:utility}. It improves spatial awareness and helps the operator to safely move away from the sediment formations toward open, accessible areas and avoid obstructing other scuba divers in the process~\cite{islam2024computer,abdullah2023caveseg}.

\vspace{1mm}
\noindent
\extension{\textbf{Operator-ROV shared autonomy}. In subsea teleoperation, collaborative decision-making frameworks split responsibilities so that humans set high-level goals while the vehicle plans and autonomously executes low-level actions~\cite{chen2025subsene}. EgoExo++ augmented visuals strengthen this ``human-in-the-loop'' paradigm by providing an interactive shared scene where operator intent (\eg, safe altitude, keep-out zones) can be directly expressed, and the ROV autonomously confirms execution. For instance, the 2.5D view helps the operator determine a safe ground clearance, which the ROV can then autonomously maintain. Additionally, higher-level human-robot interactions can be integrated in the EgoExo++ interface: the operator can draw an intended path directly on the exocentric view~\cite{lee2016development}, then the ROV can plan and follow an optimal path accordingly. Beyond navigation, augmented visuals have high demand in shared telemanipulation tasks such as delicate object grasping~\cite{chapman2008virtual}, valve control~\cite{zhang2024adaptive}, artifact collection~\cite{bruno2018augmented}, \textit{etc}. Our proposed $360\degree$ views provide the operator with critical situational cues in such tasks. For instance, a side-view perspective helps position the ROV with respect to the target, then the operator can switch to a close-up egocentric view for precise manipulation~\cite{chen2025subsene,phung2024shared}. Overall, EgoExo++ serves as a shared perceptual layer: it enhances situational awareness with explicit geometric cues and allows the operator to specify intent more precisely, promising a safer and more effective teleoperation and telemanipulation.}

\vspace{1mm}
\noindent
\extension{\textbf{Digital Twins and shadows}.
A digital shadow creates a virtual replica of the robot and its environment, enabling operators to practice missions, rehearse manipulation tasks, and refine control strategies~\cite{jones2020characterising}. While a shadow is a passive replica, a digital twin offers a bidirectional data pipeline and predictive simulation, thereby closing the feedback loop~\cite{sjarov2020digital}. In this context, EgoExo++ complements DT-based training by providing geometrically consistent exocentric perspectives and interactive \textit{Real2Sim} 2.5D reconstructions derived from mission SLAM data. During rehearsal, such views allow operators to anticipate spatial challenges, practice navigation in cluttered cave-like terrains, and visualize manipulators reaching the target from third-person perspectives~\cite{lim2022real2sim2real}. EgoExo++ views can also be rendered on HMIs, where identical head motions may be mapped to different outcomes depending on the selected visualization mode~\cite{candeloro2015hmd}. For instance, a head tilt in egocentric mode can directly control the ROV body, whereas the same movement in exocentric mode can control the virtual camera viewpoint, with no impact on the ROV. Rehearsing such multi-visual feedback and control mappings in high-fidelity simulator engines will significantly improve operator skills in high-risk, time-critical missions.}

\begin{figure}[t]
% \vspace{-2mm}
\centering
\includegraphics[width=\linewidth]{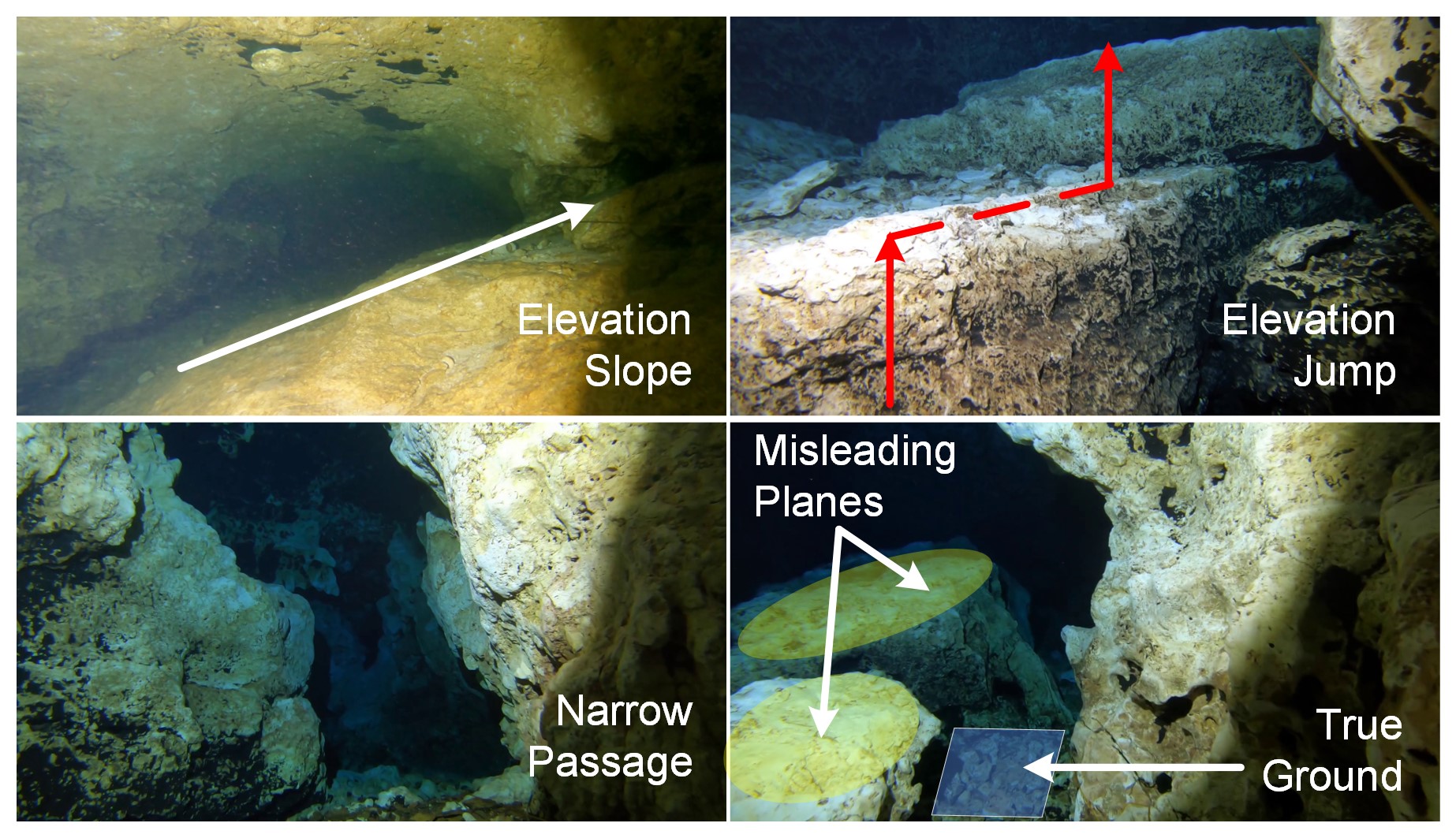}%
\vspace{-1mm}
\caption{\extension{Challenges in estimating uneven ground surface in unstructured environments: terrain complexities, such as elevation changes, narrow passages, and misleading planar obstacles, hinder the accurate ground surface reconstruction.}
}
\label{fig:ground_challenges}
\vspace{-3mm}
\end{figure}

\vspace{1mm}
\noindent
\extension{\textbf{Challenges in ground estimation}. Our field trials in diverse underwater caves and grotto systems reveal that the irregular and deceptive terrain poses several unique challenges in estimating the ground surface. As illustrated in Figure~\ref{fig:ground_challenges}, elevation slope or sharp jumps may fall outside the fitting capability of plane detection algorithms, leading to fragmented or distorted ground estimation. Additionally, narrow passages create occlusions and limited visibility of the ground, causing gaps in both feature tracking and surface mapping. The presence of large protruding structures that appear ground-like in texture can lead to incorrect segmentation of the actual ground surface. These issues collectively challenge our EgoExo++ pipeline, occasionally resulting in an unreliable representation of the terrain.}

\begin{figure}[h]
    \centering
    % \vspace{-2mm}
    \includegraphics[width=\linewidth]{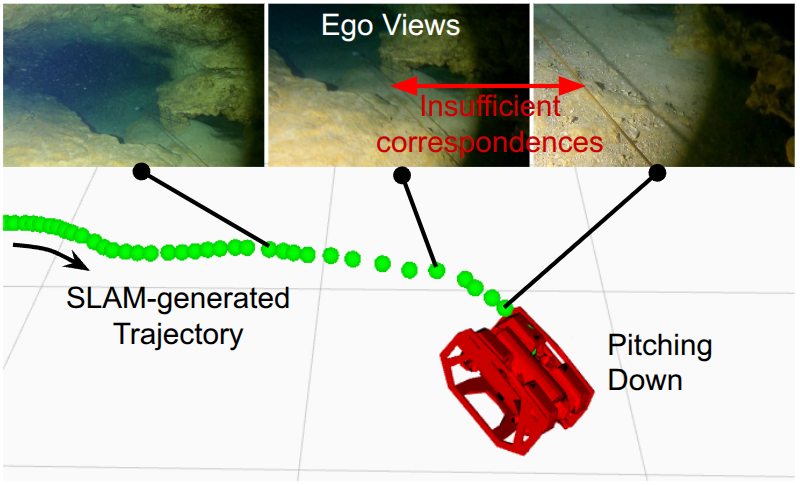}%
    \vspace{-2mm}
    \caption{An example of SLAM failure during aggressive pitch maneuvering is illustrated. A ``nose-down'' dive drives the ROV too close to the cave floor, causing the scene to become feature-deprived and leading to a SLAM tracking loss.}
    \vspace{-3mm}
    \label{fig:slam_fail}
\end{figure}

\vspace{1mm}
\noindent
\rebuttal{\textbf{SLAM dependency and failure statistics}. Since the EgoExo++ framework depends on the pose and visual features estimated by SLAM, the pose uncertainty and noisy visual conditions affect EgoExo view generation and EgoExo++ 2.5D ground reconstruction. In poor visual conditions or during abrupt motion, SLAM may temporarily lose tracking, in which case our method suspends map generation and waits for successful relocalization. Importantly, the historical ground-referenced trajectory map remains visible and provides operators with an intuitive sense of ``where am I?'', which helps them steer toward previously mapped, visually richer regions where relocalization is more likely. Overall, EgoExo++ transforms SLAM-derived data into an interactive representation that bridges egocentric and exocentric awareness, allowing operators to reason about both local maneuvering and global mission context.} 

\rebuttal{However, successful relocalization remains the responsibility of the operator and the SLAM system; EgoExo++ only supports this process passively through improved situational awareness. \finalrev{It operates as a lightweight visualization layer that deterministically consumes SLAM pose and sparse map without modeling or propagating uncertainty. This design choice prioritizes real-time operation and simplicity of integration with existing teleoperation engines.}} 

\rebuttal{Across more than $1500$ mission segments collected from our field trials in three different cave systems, SLAM tracking loss occurred in less than $5$\% of the segments, with successful relocalization achieved in roughly half of these events. The major causes are abrupt vehicle motion or collisions that drive the ROV too close to an obstacle, producing feature-deprived imagery and disrupting tracking; an example is illustrated in Figure~\ref{fig:slam_fail}. Across datasets, the terrain estimation method achieves an average success rate of approximately $92\%$ (see Table~\ref{tab:ground_plane_eval}), implying that $8\%$ of segments are rejected due to insufficient visual support. In such cases, EgoExo++ omits unreliable ground patches and leaves a hole in the reconstructed scene rather than rendering incorrect geometry.
}
\section{Conclusion and Future Work}
% \vspace{-1mm}
This work presents an AR-based framework to synthesize exocentric camera views from egocentric feed in real-time for improved underwater ROV teleoperation. A pose geometry-based closed-form solution is formulated for the proposed EgoExo++ problem and then integrated into a visual SLAM backbone. \extension{The end-to-end pipeline only requires a sequence of past egocentric views to generate 2D/2.5D exocentric views with the accurate ROV model projected onto them.} The proof-of-concept is validated by ground plane estimation and reprojection error analyses in a series of 2D indoor navigation experiments. \extension{Subsequent field experiments are conducted to demonstrate the effectiveness of 2.5D scene rendering in unstructured underwater cave scenarios.} \rebuttal{We validate the system through two subjective studies: one in simulation and one using real underwater datasets. These studies demonstrate improved system usability (SUS), reduced perceived workload (NASA-TLX), and quantitative gains in teleoperation performance, including more efficient paths and faster mission completion compared to egocentric-only baselines.} We are currently exploring more comprehensive multi-sensor fusion-based underwater SLAM backbones, such as the SVIn2~\cite{RahmanIJRR2022}, for more accurate and robust estimation. \rebuttal{We will further extend our simulation platform with interactive tools to enable advanced teleoperation studies such as confined-space navigation, SLAM-recovery behavior, close-up inspection, complex maneuvering around structures, etc.}

\section*{Acknowledgments}
\vspace{-1mm}
This work is supported in part by the NSF grants $2330416$, $2534503$, and $2545370$. The authors would like to acknowledge the help from Woodville Karst Plain Project (WKPP), El Centro Investigador del Sistema Acuífero de Quintana Roo A.C. (CINDAQ),  Global Underwater Explorers (GUE), Ricardo Constantino, and Project Baseline in providing access to challenging underwater caves. The authors are also grateful for equipment support by Halcyon Dive Systems, Teledyne FLIR LLC, and KELDAN GmbH lights. We also appreciate the participants of our user study for their time and valuable feedback during the evaluation process. Finally, we thank the anonymous reviewers for their insightful suggestions, which significantly improved this paper.

% \printbibliography
\bibliographystyle{ieeetr}
\bibliography{references,VIO_SLAM_Caves,robopi_pubs,afrl_pubs}

@article{ORBSLAM3_TRO,
  title={{ORB-SLAM3: An Accurate Open-Source Library for Visual, Visual-Inertial and Multi-Map SLAM}},
  author={Campos, Carlos and Elvira, Richard and G\'omez, Juan J. and Montiel, Jos\'e M. M. and Tard\'os, Juan D.},
  journal={IEEE Transactions on Robotics}, 
  volume={37},
  number={6},
  pages={1874-1890},
  year={2021}
 }

@article{macario2022comprehensive,
  title={{A Comprehensive Survey of Visual SLAM Algorithms}},
  author={Macario Barros, Andr{\'e}a and Michel, Maugan and Moline, Yoann and Corre, Gwenol{\'e} and Carrel, Fr{\'e}d{\'e}rick},
  journal={Robotics},
  volume={11},
  number={1},
  pages={24},
  year={2022},
  publisher={MDPI}
}

@article{kazerouni2022survey,
  title={{A Survey of State-of-the-art on Visual SLAM}},
  author={Kazerouni, Iman Abaspur and Fitzgerald, Luke and Dooly, Gerard and Toal, Daniel},
  journal={Expert Systems with Applications},
  volume={205},
  pages={117734},
  year={2022},
  publisher={Elsevier}
}

@article{RahmanIJRR2022,
  author = {Sharmin Rahman and Alberto {Quattrini Li}  and Ioannis Rekleitis},
  title = {{SVIn2: A Multi-sensor Fusion-based Underwater SLAM System}},
  journal = {International Journal of Robotics Research},
  year={2022},
  month={July},
  volume = {41},
  number = {11-12},
  pages = {1022-1042},
  label ={J17},
  category={Journal Papers}
}

@InProceedings{ChatzispyrouICRA2026,
  author = 	 {Michalis Chatzispyrou and Luke Horgan and Hyunkil Hwang and Harish Sathishchandra and Chinmay Burgul and Monika Roznere and Alberto {Quattrini Li} and Philippos Mordohai and Ioannis Rekleitis},
  title = 	 {{Mapping Pamir: Multi-Session Visual/Inertial SLAM and 3D Reconstruction of an Underwater Shipwreck}},
  booktitle = {IEEE International Conference on Robotics and Automation (ICRA)},
  year = 	 2026,
  month = 	 {Jun.},
  address = 	 {Vienna, Austria},
  note = 	 {(accepted)}
}

@InProceedings{MohammadiICMLA2023,
  author = 	 {Mohammadreza Mohammadi and Sheng-En Huang and Titon Barua and Ioannis Rekleitis and Md Jahidul Islam and Ramtin Zand},
  title = 	 {{Caveline Detection at the Edge for Autonomous Underwater Cave Exploration and Mapping}},
  booktitle = {IEEE International Conference on Machine Learning and Applications (ICMLA)},
  pages={1392-1398},
  year = 	 2023,
  month = 	 {Dec.},
  address = 	 {Jacksonville, FL, USA},
}

@InProceedings{JoshiIROS2019,
  author = 	 {Bharat Joshi and Sharmin Rahman and Michail Kalaitzakis and Brennan Cain and James Johnson and Marios Xanthidis and Nare Karapetyan and Alan Hernandez and Alberto {Quattrini Li} and Nikolaos Vitzilaios and Ioannis Rekleitis},
  title = 	 {{Experimental Comparison of Open Source Visual-Inertial-Based State Estimation Algorithms in the Underwater Domain}},
  booktitle = {IEEE/RSJ International Conference on Intelligent Robots and Systems (IROS)},
  year = 	 2019,
  pages = 	 {7221--7227},
  month = 	 {Nov.},
}

@InProceedings{JoshiAUV2022,
  author = 	 {Bharat Joshi and Marios Xanthidis and Monika Roznere and Nathaniel J. Burgdorfer and Philippos Mordohai and Alberto {Quattrini Li} and Ioannis Rekleitis},
  title = 	 {{Underwater Exploration and Mapping}},
  booktitle = {IEEE OES AUV Symposium},
  year = 	 2022,
  pages = 	 {1-7},
  month = 	 {Sept.},
  address = 	 {Singapore},
  label ={N31},
  category = {Editorials; Lightly Refereed Conference and Workshop Papers}
}

@article{chatzispyrou2025mapping,
  title={{Mapping the Catacombs: An Underwater Cave Segment of the Devil's Eye System}},
  author={Chatzispyrou, Michalis and Horgan, Luke and Hwang, Hyunkil and Sathishchandra, Harish and Roznere, Monika and {Quattrini Li}, Alberto  and Mordohai, Philippos and Rekleitis, Ioannis},
  journal={ArXiv preprint arXiv:2507.06397},
  year={2025}
}

@inproceedings{lago2024visual,
  title={Visual-inertial odometry for metric-scale mapping of underwater caves},
  author={Lago, Arthur and Neves, Nuno and Ventura, Rodrigo and Gaspar, Jos{\'e}},
  booktitle={2024 IEEE International Conference on Autonomous Robot Systems and Competitions (ICARSC)},
  pages={183--188},
  year={2024},
  organization={IEEE}
}

@article{siegel2023robotic,
  title={{Robotic Survey and 3-D Mapping of Underwater Caves using a SUNFISH{\textregistered} Autonomous Underwater Vehicle}},
  author={Siegel, VL and Stone, WC and Richmond, K},
  journal={LPI Contributions},
  volume={2697},
  pages={1037},
  year={2023}
}

@inproceedings{lensgraf2023buoyancy,
  title={{Buoyancy Enabled Autonomous Underwater Construction with Cement Blocks}},
  author={Lensgraf, Samuel and Balkcom, Devin and {Quattrini Li}, Alberto },
  booktitle={IEEE International Conference on Robotics and Automation (ICRA)},
  pages={5207--5213},
  year={2023}
}

@article{cai2020three,
  title={{Three-dimensional Obstacle Avoidance for Autonomous Underwater Robot}},
  author={Cai, Wenyu and Wu, Yan and Zhang, Meiyan},
  journal={IEEE Sensors Letters},
  volume={4},
  number={11},
  pages={1--4},
  year={2020},
  publisher={IEEE}
}

@article{palomeras2019autonomous,
  title={{Autonomous Exploration of Complex Underwater Environments using a Probabilistic Next-best-view Planner}},
  author={Palomeras, Narc{\'\i}s and Hurt{\'o}s, Natalia and Vidal, Eduard and Carreras, Marc},
  journal={IEEE Robotics and Automation Letters},
  volume={4},
  number={2},
  pages={1619--1625},
  year={2019},
  publisher={IEEE}
}

@inproceedings{ferland2009egocentric,
  title={{Egocentric and Exocentric Teleoperation Interface using Real-time, 3D Video Projection}},
  author={Ferland, Fran{\c{c}}ois and Pomerleau, Fran{\c{c}}ois and Le Dinh, Chon Tam and Michaud, Fran{\c{c}}ois},
  booktitle={Proceedings of the 4th ACM/IEEE International Conference on Human Robot Interaction},
  pages={37--44},
  year={2009}
}

@inproceedings{xia2022virtual,
  title={{Virtual Telepresence for the Future of ROV Teleoperations: Opportunities and Challenges}},
  author={Xia, Pengxiang and McSweeney, Kevin and Wen, Feng and Song, Zhuoyuan and Krieg, Michael and Li, Shuai and Yu, Xiao and Crippen, Kent and Adams, Jonathan and Du, Eric Jing},
  booktitle={SNAME Offshore Symposium},
  pages={D011S001R001},
  year={2022},
  organization={SNAME}
}

@inproceedings{xia2023rov,
  title={{ROV Teleoperation based on Sensory Augmentation and Digital Twins}},
  author={Xia, Pengxiang and McSweeney, Kevin P and Song, Zhuoyuan and Du, Eric},
  booktitle={Offshore Technology Conference},
  pages={D031S041R004},
  year={2023},
  organization={OTC}
}

@article{zhou2023embodied,
  title={{Embodied Robot Teleoperation based on High-Fidelity Visual-Haptic Simulator: Pipe-Fitting Example}},
  author={Zhou, Tianyu and Xia, Pengxiang and Ye, Yang and Du, Jing},
  journal={Journal of Construction Engineering and Management},
  volume={149},
  number={12},
  pages={04023129},
  year={2023},
  publisher={American Society of Civil Engineers}
}

@article{xia2023visual,
  title={{Visual-Haptic Feedback for ROV Subsea Navigation Control}},
  author={Xia, Pengxiang and You, Hengxu and Du, Jing},
  journal={Automation in Construction},
  volume={154},
  pages={104987},
  year={2023},
  publisher={Elsevier}
}

@article{bruno2018augmented,
  title={{Augmented Reality Visualization of Scene Depth for Aiding ROV Pilots in Underwater Manipulation}},
  author={Bruno, Fabio and Lagudi, Antonio and Barbieri, Loris and Rizzo, Domenico and Muzzupappa, Maurizio and De Napoli, Luigi},
  journal={Ocean Engineering},
  volume={168},
  pages={140--154},
  year={2018},
  publisher={Elsevier}
}

@article{girbes2020haptic,
  title={{Haptic and Visual Feedback Assistance for Dual-Arm Robot Teleoperation in Surface Conditioning Tasks}},
  author={Girbes-Juan, Vicent and Schettino, Vinicius and Demiris, Yiannis and Tornero, Josep},
  journal={IEEE Transactions on Haptics},
  volume={14},
  number={1},
  pages={44--56},
  year={2020},
  publisher={IEEE}
}

@inproceedings{konoplin2019development,
  title={{Development of Intellectual Support System for ROV Operators}},
  author={Konoplin, A Yu and Konoplin, N Yu and Filaretov, VF},
  booktitle={IOP Conference Series: Earth and Environmental Science},
  volume={272},
  pages={032101},
  year={2019},
  organization={IOP Publishing}
}

@article{wishnak2022new,
  title={{New Frontiers in Ocean Exploration: The Ocean Exploration Trust}},
  author={Wishnak, Samantha},
  journal={NOAA Ocean Exploration, and Schmidt Ocean Institute 2021 Field Season},
  year={2022}
}

@article{buzzacott2009american,
  title={{American Cave Diving Fatalities 1969-2007}},
  author={Buzzacott, Peter L and Zeigler, Erin and Denoble, Petar and Vann, Richard},
  journal={{International Journal of Aquatic Research and Education}},
  volume={3},
  number={2},
  pages={7},
  year={2009}
}

@inproceedings{olson2011apriltag,
  title={{AprilTag: A Robust and Flexible Visual Fiducial System}},
  author={Olson, Edwin},
  booktitle={2011 IEEE International Conference on Robotics and Automation},
  pages={3400--3407},
  year={2011},
  organization={IEEE}
}

@article{kennedy2019unknown,
  title={{The Unknown and the Unexplored: Insights into the Pacific Deep-sea Following NOAA CAPSTONE Expeditions}},
  author={Kennedy, Brian RC and Cantwell, Kasey and Malik, Mashkoor and Kelley, Christopher and Potter, Jeremy and Elliott, Kelley and Lobecker, Elizabeth and Gray, Lindsay McKenna and Sowers, Derek and White, Michael P and France, Scott C. and Auscavitch, Steven and Mah, Christopher and Moriwake, Virginia and Bingo, Sarah RD and Putts, Meagan and Rotjan, Randi D. },
  journal={Frontiers in Marine Science},
  volume={6},
  pages={480},
  year={2019},
  publisher={Frontiers Media SA}
}

@article{rumson2021application,
  title={{The Application of Fully Unmanned Robotic Systems for Inspection of Subsea Pipelines}},
  author={Rumson, Alexander G},
  journal={Ocean Engineering},
  volume={235},
  pages={109214},
  year={2021},
  publisher={Elsevier}
}

@article{manjunatha2018low,
  title={{A Low Cost Underwater Robot with Grippers for Visual Inspection of External Pipeline Surface}},
  author={Manjunatha, M and Selvakumar, A Arockia and Godeswar, Vivek P and Manimaran, R},
  journal={Procedia Computer Science},
  volume={133},
  pages={108--115},
  year={2018},
  publisher={Elsevier}
}

@inproceedings{elor2021catching,
  title={{Catching Jellies in Immersive Virtual Reality: A Comparative Teleoperation Study of ROVs in Underwater Capture Tasks}},
  author={Elor, Aviv and Thang, Tiffany and Hughes, Benjamin Paul and Crosby, Alison and Phung, Amy and Gonzalez, Everardo and Katija, Kakani and Haddock, Steven HD and Martin, Eric J and Erwin, Benjamin Eric and Takayama, Leila},
  booktitle={Proceedings of the 27th ACM Symposium on Virtual Reality Software and Technology},
  pages={1--10},
  year={2021}
}

@article{jin2022hovering,
  title={{Hovering Control of UUV through Underwater Object Detection Based on Deep Learning}},
  author={Jin, Han-Sol and Cho, Hyunjoon and Jiafeng, Huang and Lee, Ji-Hyeong and Kim, Myung-Jun and Jeong, Sang-Ki and Ji, Dae-Hyeong and Joo, Kibum and Jung, Dongwook and Choi, Hyeung-Sik},
  journal={Ocean Engineering},
  volume={253},
  pages={111321},
  year={2022},
  publisher={Elsevier}
}

@INPROCEEDINGS{gawel2018aerial,
  author={Gawel, Abel and Lin, Yukai and Koutros, Théodore and Siegwart, Roland and Cadena, Cesar},
  booktitle={2018 IEEE International Symposium on Safety, Security, and Rescue Robotics (SSRR)}, 
  title={{Aerial-Ground Collaborative Sensing: Third-Person View for Teleoperation}}, 
  year={2018},
  volume={},
  number={},
  pages={1-7},
}

@ARTICLE{jangir2022thirdpersonmanip,
  author={Jangir, Rishabh and Hansen, Nicklas and Ghosal, Sambaran and Jain, Mohit and Wang, Xiaolong},
  journal={IEEE Robotics and Automation Letters}, 
  title={{Look Closer: Bridging Egocentric and Third-Person Views With Transformers for Robotic Manipulation}}, 
  year={2022},
  volume={7},
  number={2},
  pages={3046-3053},
}

@INPROCEEDINGS{saakes2013flyingcameras,
  author={Saakes, Daniel and Choudhary, Vipul and Sakamoto, Daisuke and Inami, Masahiko and Lgarashi, Takeo},
  booktitle={2013 23rd International Conference on Artificial Reality and Telexistence (ICAT)}, 
  title={{A Teleoperating Interface for Ground Vehicles using Autonomous Flying Cameras}}, 
  year={2013},
  volume={},
  number={},
  pages={13-19},
}

@inproceedings{inoue2023birdviewar,
author = {Inoue, Maakito and Takashima, Kazuki and Fujita, Kazuyuki and Kitamura, Yoshifumi},
title = {{BirdViewAR: Surroundings-aware Remote Drone Piloting Using an Augmented Third-person Perspective}},
year = {2023},
booktitle = {Proceedings of the 2023 CHI Conference on Human Factors in Computing Systems},
articleno = {31},
pages = {1-19},
}

@INPROCEEDINGS{nagatani2011quince,
  author={Nagatani, Keiji and Kiribayashi, Seiga and Okada, Yoshito and Tadokoro, Satoshi and Nishimura, Takeshi and Yoshida, Tomoaki and Koyanagi, Eiji and Hada, Yasushi},
  booktitle={2011 IEEE International Symposium on Safety, Security, and Rescue Robotics}, 
  title={{Redesign of Rescue Mobile Robot Quince}}, 
  year={2011},
  volume={},
  number={},
  pages={13-18},
}

@ARTICLE{livatino2021intuitiverobotteleop,
  author={Livatino, Salvatore and Guastella, Dario C. and Muscato, Giovanni and Rinaldi, Vincenzo and Cantelli, Luciano and Melita, Carmelo D. and Caniglia, Alessandro and Mazza, Riccardo and Padula, Gianluca},
  journal={IEEE Access}, 
  title={{Intuitive Robot Teleoperation Through Multi-Sensor Informed Mixed Reality Visual Aids}}, 
  year={2021},
  volume={9},
  number={},
  pages={25795-25808},
}

@article{hing2010chaseview,
author = {Hing, James and Sevcik, Keith and Oh, Paul},
year = {2010},
month = {08},
pages = {485-503},
title = {{Development and Evaluation of a Chase View for UAV Operations in Cluttered Environments}},
volume = {57},
isbn = {978-90-481-8763-8},
journal = {Journal of Intelligent and Robotic Systems},
}

@INPROCEEDINGS{sato2013spatiotemporal,
  author={Sato, Takaaki and Moro, Alessandro and Sugahara, Atsushi and Tasaki, Tsuyoshi and Yamashita, Atsushi and Asama, Hajime},
  booktitle={Proceedings of the 2013 IEEE/SICE International Symposium on System Integration}, 
  title={{Spatio-temporal Bird's-eye View Images using Multiple Fish-eye Cameras}}, 
  year={2013},
  volume={},
  number={},
  pages={753-758},
}

@ARTICLE{woods2004humanrobotcoordination,
  author={Woods, D.D. and Tittle, J. and Feil, M. and Roesler, A.},
  journal={IEEE Transactions on Systems, Man, and Cybernetics, Part C (Applications and Reviews)}, 
  title={{Envisioning Human-robot Coordination in Future Operations}}, 
  year={2004},
  volume={34},
  number={2},
  pages={210-218},
}

@ARTICLE{casper2003wtc,
  author={Casper, J. and Murphy, R.R.},
  journal={IEEE Transactions on Systems, Man, and Cybernetics, Part B (Cybernetics)}, 
  title={{Human-robot Interactions during the Robot-assisted Urban Search and Rescue Response at the World Trade Center}}, 
  year={2003},
  volume={33},
  number={3},
  pages={367-385},
}

@inproceedings{ito2008teleoperation,
  title={{A Teleoperation Interface using Past Images for Outdoor Environment}},
  author={Ito, Masataka and Sato, Noritaka and Sugimoto, Maki and Shiroma, Naoji and Inami, Masahiko and Matsuno, Fumitoshi},
  booktitle={2008 SICE Annual Conference},
  pages={3372--3375},
  year={2008},
  organization={IEEE}
}

@article{lee1980two,
  title={{Two Algorithms for Constructing A Delaunay Triangulation}},
  author={Lee, Der-Tsai and Schachter, Bruce J},
  journal={International Journal of Computer \& Information Sciences},
  volume={9},
  number={3},
  pages={219--242},
  year={1980},
  publisher={Springer}
}

@inproceedings{candeloro2015hmd,
  title={{HMD as a New Tool for Telepresence in Underwater Operations and Closed-loop Control of ROVs}},
  author={Candeloro, Mauro and Valle, Eirik and Miyazaki, Michel R and Skjetne, Roger and Ludvigsen, Martin and S{\o}rensen, Asgeir J},
  booktitle={OCEANS 2015-MTS/IEEE Washington},
  pages={1--8},
  year={2015},
  organization={IEEE}
}

@article{brooke1996sus,
  title={{SUS- A Quick and Dirty Usability Scale}},
  author={Brooke, John},
  journal={Usability Evaluation in Industry},
  volume={189},
  number={194},
  pages={4--7},
  year={1996},
  publisher={London, England}
}

@inproceedings{murata2014teleoperation,
  title={{Teleoperation System using Past Image Records for Mobile Manipulator}},
  author={Murata, Ryosuke and Songtong, Sira and Mizumoto, Hisashi and Kon, Kazuyuki and Matsuno, Fumitoshi},
  booktitle={IEEE/RSJ International Conference on Intelligent Robots and Systems},
  pages={4340-4345},
  year={2014}
}

@article{thomason2019comparison,
  title={{A Comparison of Adaptive View Techniques for Exploratory 3D Drone Teleoperation}},
  author={Thomason, John and Ratsamee, Photchara and Orlosky, Jason and Kiyokawa, Kiyoshi and Mashita, Tomohiro and Uranishi, Yuki and Takemura, Haruo},
  journal={ACM Transactions on Interactive Intelligent Systems},
  volume={9},
  pages={1--19},
  year={2019}
}

@article{erat2018drone,
  title={{Drone-augmented Human Vision: Exocentric Control for Drones Exploring Hidden Areas}},
  author={Erat, Okan and Isop, Werner Alexander and Kalkofen, Denis and Schmalstieg, Dieter},
  journal={IEEE Transactions on Visualization and Computer Graphics},
  volume={24},
  number={4},
  pages={1437--1446},
  year={2018}
}

@inproceedings{shiroma2004study,
  title={{Study on Effective Camera Images for Mobile Robot Teleoperation}},
  author={Shiroma, Naoji and Sato, Noritaka and Chiu, Yu-huan and Matsuno, Fumitoshi},
  booktitle={IEEE International Workshop on Robot and Human Interactive Communication (RO-MAN)},
  pages={107--112},
  year={2004}
}

@article{zollmann2014flyar,
  title={{FlyAR: Augmented Reality Supported Micro Aerial Vehicle Navigation}},
  author={Zollmann, Stefanie and Hoppe, Christof and Langlotz, Tobias and Reitmayr, Gerhard},
  journal={IEEE Transactions on Visualization and Computer Graphics},
  volume={20},
  number={4},
  pages={560--568},
  year={2014},
  publisher={IEEE}
}

@inproceedings{lee2016development,
  title={{Development of the Human Interactive Autonomy for the Shared Teleoperation of Mobile Robots}},
  author={Lee, Kwang-Hyun and Mehmood, Usman and Ryu, Jee-Hwan},
  booktitle={2016 IEEE/RSJ International Conference on Intelligent Robots and Systems (IROS)},
  pages={1524--1529},
  year={2016},
  organization={IEEE}
}

@misc{eustice2005large,
  author = {Eustice, Ryan M.},
  title  = {Large-Area Visually Augmented Navigation for Autonomous Underwater Vehicles},
  year   = {2005},
  note   = {PhD Thesis, Massachusetts Institute of Technology}
}

@article{phung2024shared,
  title={{A Shared Autonomy System for Precise and Efficient Remote Underwater Manipulation}},
  author={Phung, Amy and Billings, Gideon and Daniele, Andrea F and Walter, Matthew R and Camilli, Richard},
  journal={IEEE Transactions on Robotics},
  year={2024},
  publisher={IEEE}
}

@article{fischler1981random,
  title={{Random Sample Consensus: A Paradigm for Model Fitting with Applications to Image Analysis and Automated Cartography}},
  author={Fischler, Martin A and Bolles, Robert C},
  journal={Communications of the ACM},
  volume={24},
  number={6},
  pages={381--395},
  year={1981},
  publisher={ACM New York, NY, USA}
}

@inproceedings{wang2019monocular,
  title={{Monocular Plan View Networks for Autonomous Driving}},
  author={Wang, Dequan and Devin, Coline and Cai, Qi-Zhi and Kr{\"a}henb{\"u}hl, Philipp and Darrell, Trevor},
  booktitle={2019 IEEE/RSJ International Conference on Intelligent Robots and Systems (IROS)},
  pages={2876--2883},
  year={2019},
  organization={IEEE}
}

@inproceedings{zhu2018generative,
  title={{Generative Adversarial Frontal View to Bird View Synthesis}},
  author={Zhu, Xinge and Yin, Zhichao and Shi, Jianping and Li, Hongsheng and Lin, Dahua},
  booktitle={2018 International Conference on 3D Vision (3DV)},
  pages={454--463},
  year={2018},
  organization={IEEE}
}

@inproceedings{abbas2019geometric,
  title={{A Geometric Approach to Obtain A Bird's Eye View From An Image}},
  author={Abbas, Syed Ammar and Zisserman, Andrew},
  booktitle={2019 IEEE/CVF International Conference on Computer Vision Workshop (ICCVW)},
  pages={4095--4104},
  year={2019},
  organization={IEEE}
}

@article{li2024bevformer,
  title={{BEVFormer: Learning Bird's-Eye-View Representation from Lidar-Camera via Spatiotemporal Transformers}},
  author={Li, Zhiqi and Wang, Wenhai and Li, Hongyang and Xie, Enze and Sima, Chonghao and Lu, Tong and Yu, Qiao and Dai, Jifeng},
  journal={IEEE Transactions on Pattern Analysis and Machine Intelligence},
  year={2024},
  publisher={IEEE}
}

@inproceedings{li2023bevdepth,
  title={{BEVDepth: Acquisition of Reliable Depth for Multi-View 3d Object Detection}},
  author={Li, Yinhao and Ge, Zheng and Yu, Guanyi and Yang, Jinrong and Wang, Zengran and Shi, Yukang and Sun, Jianjian and Li, Zeming},
  booktitle={Proceedings of the AAAI Conference on Artificial Intelligence},
  volume={37},
  pages={1477--1485},
  year={2023}
}

@inproceedings{samani2023f2bev,
  title={{F2BEV: Bird's Eye View Generation from Surround-View Fisheye Camera Images for Automated Driving}},
  author={Samani, Ekta U and Tao, Feng and Dasari, Harshavardhan R and Ding, Sihao and Banerjee, Ashis G},
  booktitle={2023 IEEE/RSJ International Conference on Intelligent Robots and Systems (IROS)},
  pages={9367--9374},
  year={2023},
  organization={IEEE}
}

@inproceedings{reiher2020sim2real,
  title={{A Sim2real Deep Learning Approach for the Transformation of Images from Multiple Vehicle-mounted Cameras to A Semantically Segmented Image in Bird’s Eye View}},
  author={Reiher, Lennart and Lampe, Bastian and Eckstein, Lutz},
  booktitle={2020 IEEE 23rd International Conference on Intelligent Transportation Systems (ITSC)},
  pages={1--7},
  year={2020},
  organization={IEEE}
}

@inproceedings{lager2018remote,
  title={{Remote Operation of Unmanned Surface Vessel through Virtual Reality-A Low Cognitive Load Approach}},
  author={Lager, M{\aa}rten and Topp, Elin A and Malec, Jacek},
  booktitle={Proceedings of the 1st International Workshop on Virtual, Augmented, and Mixed Reality for HRI (VAM-HRI)},
  year={2018}
}

@article{nguyen2001virtual,
  title={{Virtual Reality Interfaces for Visualization and Control of Remote Vehicles}},
  author={Nguyen, Laurent A and Bualat, Maria and Edwards, Laurence J and Flueckiger, Lorenzo and Neveu, Charles and Schwehr, Kurt and Wagner, Michael D and Zbinden, Eric},
  journal={Autonomous Robots},
  volume={11},
  pages={59--68},
  year={2001},
  publisher={Springer}
}

@inproceedings{okura2013teleoperation,
  title={{Teleoperation of Mobile Robots by Generating Augmented Free-Viewpoint Images}},
  author={Okura, Fumio and Ueda, Yuko and Sato, Tomokazu and Yokoya, Naokazu},
  booktitle={2013 IEEE/RSJ International Conference on Intelligent Robots and Systems},
  pages={665--671},
  year={2013},
  organization={IEEE}
}

@article{luo2024intention,
  title={{Intention-driven Ego-to-Exo Video Generation}},
  author={Luo, Hongchen and Zhu, Kai and Zhai, Wei and Cao, Yang},
  journal={ArXiv Preprint arXiv:2403.09194},
  year={2024}
}

@article{xia2023sensory,
  title={{Sensory Augmentation for Subsea Robot Teleoperation}},
  author={Xia, Pengxiang and Xu, Fang and Song, Zhuoyuan and Li, Shuai and Du, Jing},
  journal={Computers in Industry},
  volume={145},
  pages={103836},
  year={2023},
  publisher={Elsevier}
}

@article{thatipelli2025egocentric,
  title={{Egocentric and Exocentric Methods: A Short Survey}},
  author={Thatipelli, Anirudh and Lo, Shao-Yuan and Roy-Chowdhury, Amit K},
  journal={Computer Vision and Image Understanding},
  pages={104371},
  year={2025},
  publisher={Elsevier}
}

@article{yoon2025learning,
  title={{Learning Viewpoint Control from Human-Initiated Transitions for Teleoperation in Construction}},
  author={Yoon, Sungboo and Park, Moonseo and Ahn, Changbum R},
  journal={Advanced Engineering Informatics},
  volume={68},
  pages={103665},
  year={2025},
  publisher={Elsevier}
}

@inproceedings{manderson2016texture,
  title={{Texture-Aware SLAM using Stereo Imagery and Inertial Information}},
  author={Manderson, Travis and Shkurti, Florian and Dudek, Gregory},
  booktitle={2016 13th Conference On Computer And Robot Vision (CRV)},
  pages={456--463},
  year={2016},
  organization={IEEE}
}

@article{zhang2024adaptive,
  title={{The Adaptive Bilateral Control of Underwater Manipulator Teleoperation System with Uncertain Parameters and External Disturbance}},
  author={Zhang, Jianjun and Xia, Manjiang and Li, Shasha and Liu, Zhiqiang and Yang, Jinxian},
  journal={Electronics},
  volume={13},
  number={6},
  pages={1122},
  year={2024},
  publisher={MDPI}
}

@article{chapman2008virtual,
  title={{Virtual Exploration of Underwater Archaeological Sites: Visualization and Interaction in Mixed Reality Environments}},
  author={Chapman, Paul and Roussel, David and Drap, Pierre and Haydar, Mohamed},
  journal={International Symposium on Virtual Reality, Archaeology and Intelligent Cultural Heritage},
  year={2008}
}

@article{jones2020characterising,
  title={{Characterising the Digital Twin: A Systematic Literature Review}},
  author={Jones, David and Snider, Chris and Nassehi, Aydin and Yon, Jason and Hicks, Ben},
  journal={CIRP Journal of Manufacturing Science and Technology},
  volume={29},
  pages={36--52},
  year={2020},
  publisher={Elsevier}
}

@inproceedings{sjarov2020digital,
  title={{The Digital Twin Concept in Industry--A Review and Systematization}},
  author={Sjarov, Martin and Lechler, Tobias and Fuchs, Jonathan and Brossog, Matthias and Selmaier, Andreas and Faltus, Florian and Donhauser, Toni and Franke, J{\"o}rg},
  booktitle={2020 25th IEEE International Conference on Emerging Technologies and Factory Automation (ETFA)},
  volume={1},
  pages={1789--1796},
  year={2020},
  organization={IEEE}
}

@inproceedings{lim2022real2sim2real,
  title={{Real2sim2real: Self-Supervised Learning of Physical Single-Step Dynamic Actions for Planar Robot Casting}},
  author={Lim, Vincent and Huang, Huang and Chen, Lawrence Yunliang and Wang, Jonathan and Ichnowski, Jeffrey and Seita, Daniel and Laskey, Michael and Goldberg, Ken},
  booktitle={2022 International Conference on Robotics and Automation (ICRA)},
  pages={8282--8289},
  year={2022},
  organization={IEEE}
}

@article{massone2024novel,
  title={{A Novel 3D Reconstruction Sensor Using a Diving Lamp and a Camera for Underwater Cave Exploration}},
  author={Massone, Quentin and Druon, S{\'e}bastien and Triboulet, Jean},
  journal={Sensors (Basel, Switzerland)},
  volume={24},
  number={12},
  pages={4024},
  year={2024}
}

@inproceedings{stewart2016interactive,
  title={{An Interactive Interface for Multi-Pilot ROV Intervention}},
  author={Stewart, Andrew and Ryden, Fredrik and Cox, Ryan},
  booktitle={OCEANS 2016-Shanghai},
  pages={1--6},
  year={2016},
  organization={IEEE}
}

@incollection{hart1988development,
  title={{Development of NASA-TLX (Task Load Index): Results of Empirical and Theoretical Research}},
  author={Hart, Sandra G and Staveland, Lowell E},
  booktitle={Advances in Psychology},
  volume={52},
  pages={139--183},
  year={1988},
  publisher={Elsevier}
}

@article{islam2024computer,
  title={{Computer Vision Applications in Underwater Robotics and Oceanography}},
  author={Islam, Md Jahidul and {Quattrini Li}, Alberto  and Girdhar, Yogesh A and Rekleitis, Ioannis},
  journal={Computer Vision: Challenges, Trends, and Opportunities},
  pages={173--196},
  year={2024},
  publisher={CRC Press}
}

@article{mo2021fast,
  title={{Fast Direct Stereo Visual SLAM}},
  author={Mo, Jiawei and Islam, Md Jahidul and Sattar, Junaed},
  journal={IEEE Robotics and Automation Letters},
  volume={7},
  number={2},
  pages={778--785},
  year={2021},
  publisher={IEEE}
}

@article{islam2021robot,
  title={{Robot-to-Robot Relative Pose Estimation using Humans as Markers}},
  author={Islam, Md Jahidul and Mo, Jiawei and Sattar, Junaed},
  journal={Autonomous Robots},
  volume={45},
  number={4},
  pages={579--593},
  year={2021},
  publisher={Springer}
}

@inproceedings{islam2024eob,
    author={Islam, Md Jahidul},
    title={{Eye on the Back: Augmented Visuals for Improved ROV Teleoperation in Deep Water Surveillance and Inspection}},
    booktitle={Ocean Sensing and Monitoring XVISPIE Defense and Commercial Sensing},
    volume={13061},
    pages={21--25},
    year={2024},
    address={Maryland, USA},
    organization={SPIE}
}

@inproceedings{yu2022udepth,    
author={Yu, Boxiao and Wu, Jiayi and Islam, Md Jahidul},    
title={{UDepth: Fast Monocular Depth Estimation for Visually-guided Underwater Robots}},    
booktitle={IEEE International Conference on Robotics and Automation (ICRA)},    
year={2023},
pages={3116-3123},
organization={IEEE}
}

@inproceedings{abdullah2023caveseg,
  title={{CaveSeg: Deep Semantic Segmentation and Scene Parsing for Autonomous Underwater Cave Exploration}},
  author={Abdullah, Adnan and Barua, Titon and Tibbetts, Reagan and Chen, Zijie and Islam, Md Jahidul and Rekleitis, Ioannis},
  booktitle={IEEE International Conference on Robotics and Automation (ICRA)},
  year={2024},
  pages={3781-3788},
}

@article{abdullah2024human,
  title={{Human-Machine Interfaces for Subsea Telerobotics: From Soda-straw to Natural Language Interactions}},
  author={Abdullah, Adnan and Blow, David and Chen, Ruo and Uthai, Thanakon and Du, Eric Jing and Islam, Md Jahidul},
  journal={{ArXiv Preprint arXiv:2412.01753}},
  year={2024}
}

@inproceedings{gupta2025demonstrating,
  title={{Demonstrating CavePI: Autonomous Exploration of Underwater Caves by Semantic Guidance}},
  author={Gupta, Alankrit and Abdullah, Adnan and Li, Xianyao and Ramesh, Vaishnav and Rekleitis, Ioannis and Islam, Md Jahidul},
  booktitle={Robotics: Science and Systems (RSS)},
  year={2025}
}

@article{abdullah2025nemesys,
  title={{NemeSys: An Online Underwater Explorer with Goal-Driven Adaptive Autonomy}},
  author={Abdullah, Adnan and Gupta, Alankrit and Ramesh, Vaishnav and Patel, Shivali and Islam, Md Jahidul},
  journal={ArXiv Preprint arXiv:2507.11889},
  year={2025}
}

@inproceedings{yu2023weakly,
  title={{Weakly Supervised Caveline Detection For AUV Navigation Inside Underwater Caves}},
  author={Yu, Boxiao and Tibbetts, Reagan and Barua, Titon and Morales, Ailani and Rekleitis, Ioannis and Islam, Md Jahidul},
  booktitle={IEEE/RSJ International Conference on Intelligent Robots and Systems (IROS)},
  pages={9933--9940},
  year={2023},
  organization={IEEE}
}

@inproceedings{wu2023sdu_sfm,
  title={{3D Reconstruction of Underwater Scenes using Nonlinear Domain Projection}},
  author={Wu, Jiayi and Yu, Boxiao and Islam, Md Jahidul},
  booktitle={2023 IEEE Conference on Artificial Intelligence (CAI)},
  pages={359--361},
  year={2023},
  organization={IEEE},
  note={Best Paper Award}
}

@inproceedings{abdullah2024ego,
  title={{Ego-to-Exo: Interfacing Third Person Visuals from Egocentric Views in Real-time for Improved ROV Teleoperation}},
  author={Abdullah, Adnan and Chen, Ruo and Rekleitis, Ioannis and Islam, Md Jahidul},
  booktitle  = {International Symposium on Robotics Research (ISRR)},
  year={2024}
}

@inproceedings{chen2025subsene,
  title={{SubSense: VR-Haptic and Motor Feedback for Immersive Control in Subsea Telerobotics}},
  author={Chen, Ruo and Blow, David and Abdullah, Adnan and Islam, Md Jahidul},
  booktitle={The OCEANS Conference},
  year={2025},
  organization={IEEE OES}
}

@inproceedings{blow2025detection,
    author={Blow, David and Abdullah, Adnan and Sheldon, Jennifer and Zhu, Weidong and Rampazzi, Sara and Islam, Md Jahidul},
    title={{Detection and Localization of Acoustic Vulnerabilities of Underwater Data Centers for Remote Surveillance}},
    booktitle={SPIE Defense and Commercial Sensing},
    year={2025},
    address={Ocean Sensing and Monitoring XVI},
    organization={SPIE}
}

@article{abdullah2025active,
  title={{Active Localization of Close-range Adversarial Acoustic Sources for Underwater Data Center Surveillance}},
  author={Abdullah, Adnan and Blow, David and Rampazzi, Sara and Islam, Md Jahidul},
  journal={Under review at the IEEE Journal of Oceanic Engineering (JOE). ArXiv:2510.20122},
  year={2025}
}

\end{document}